\documentclass[conference]{IEEEtran}

\usepackage[noadjust]{cite}
\usepackage{multicol}
\usepackage{booktabs}
\usepackage{colortbl}
\ifCLASSINFOpdf
   \usepackage[pdftex]{graphicx}
   \graphicspath{{../pdf/}{../png/}}

\usepackage{subcaption,graphicx,url,times,multirow,amsmath,amssymb,algorithm,xspace,epsfig,todonotes,array,caption,color}

\usepackage{float}
\usepackage[noend]{algpseudocode}

\definecolor{red}{RGB}{255,0,0}

\title{Evaluating Generalisation in \\General Video Game Playing}
\author{
\IEEEauthorblockN{Martin Balla}
\IEEEauthorblockA{Queen Mary University of London\\
m.balla@qmul.ac.uk}
\and
\IEEEauthorblockN{Simon M. Lucas}
\IEEEauthorblockA{Queen Mary University of London\\
simon.lucas@qmul.ac.uk}
\and
\IEEEauthorblockN{Diego Perez-Liebana}
\IEEEauthorblockA{Queen Mary University of London\\
diego.perez@qmul.ac.uk}}

\IEEEoverridecommandlockouts

\IEEEpubid{\begin{minipage}{\textwidth}\ \\[12pt]
978-1-7281-4533-4/20/\$31.00 \copyright 2020 IEEE
\end{minipage}}

\begin{document}

\maketitle

\begin{abstract}
The General Video Game Artificial Intelligence (GVGAI) competition has been running for several years with various tracks. This paper focuses on the challenge of the GVGAI learning track in which 3 games are selected and 2 levels are given for training, while 3 hidden levels are left for evaluation. This setup poses a difficult challenge for current Reinforcement Learning (RL) algorithms, as they typically require much more data. This work investigates 3 versions of the Advantage Actor-Critic (A2C) algorithm trained on a maximum of $2$ levels from the available $5$ from the GVGAI framework and compares their performance on all levels. The selected sub-set of games have different characteristics, like stochasticity, reward distribution and objectives. We found that stochasticity improves the generalisation, but too much can cause the algorithms to fail to learn the training levels. The quality of the training levels also matters, different sets of training levels can boost generalisation over all levels. In the GVGAI competition agents are scored based on their win rates and then their scores achieved in the games. We found that solely using the rewards provided by the game might not encourage winning.
\end{abstract}

\section{Introduction}
Games are commonly used to benchmark Reinforcement Learning (RL) algorithms, but in many cases training is performed on a single level, which is also used for evaluation. In this work, the General Video Game AI (GVGAI) framework has been used as it provides a large number of games with at least $5$ levels for each one of them. The GVGAI Learning competition has been running for a few years and received several entries, but due to the difficulty of the games and the requirement to generalise to new levels to perform well on the evaluations, even the best agents do not perform better than the random agent during evaluation. This paper investigates the possibility of generalizing to new levels only using a maximum of $2$ levels for training.

In the majority of Deep Reinforcement Learning experiments, the training domain is used to test the performance of the algorithm. However, this approach does not give enough information about the agent's real capabilities. Often agents just memorise scenarios, learning a sequence of actions, instead of showing intelligent and adaptable behaviours. These behaviours might not be obvious if there are no test levels, as they might just overfit to the set of training levels. The GVGAI framework provides multiple levels (which have the same rules, but some variants on the map layout or sprite types) for each game, so it provides a great testing opportunity for RL agents. Due to the Video Game Description Language (VGDL) used by GVGAI to build games, it is easy to create new games and levels or even procedurally generate levels during training~\cite{justesen2018illuminating}. Generalisation recently received more attention for benchmarking RL agents with challenges like Sonic Retro~\cite{nichol2018gotta} and Obstacle Tower~\cite{juliani2019obstacle}.

The main objective of this paper is to investigate to which extent generalisation happens in Reinforcement Learning, when the agent has a very limited number of levels for training. We trained 3 different version of the A2C algorithm on 4 games on various training levels and compared them on all available levels of that game. Some sets of training levels have been evaluated in order to investigate if having training levels, that are more representative of the different features of the game can improve generalisation.

This paper introduces GVGAI in Section~\ref{sec:gvgai}, to then describe the most relevant work around Deep Reinforcement Learning, generalisation in reinforcement learning and how has this been explored in GVGAI in Section~\ref{sec:lit}. Section~\ref{sec:methods} describes the methods used in this study. Section~\ref{sec:expsetup} outlines the experiments performed and the results observed are detailed in Section~\ref{sec:res}. Finally, Section~\ref{sec:conc} concludes the paper with the main takeaways and future work.

\section{The General Video Game AI Framework} \label{sec:gvgai}

The General Video Game AI Framework (GVGAI) framework is a Java benchmark, evolved from the original py-vgdl implementation by Tom Schaul~\cite{schaul13}. It is developed with the objective of centering research in general video game playing. Rather than focusing on a single game, for which ad-hoc heuristics can be built, the GVGAI framework proposes planning and learning challenges for playing agents in multiple games. The GVGAI competition~\cite{perez20162014}, run since 2014 and built around this benchmark, has helped the framework reach a widespread usage in research and education~\cite{perez2019general}.

At the time of writing, GVGAI has a collection of more than $180$ two-dimensional arcade-like games. These games are described using the Video Game Description Language (VGDL;~\cite{schaul13}), which allows the definition of games and levels in a text-based format. Games are designed around a collection of sprites, which have their own behaviours and properties. These sprites can interact with each other by means of collisions, and the effects of these interactions can determine the winning condition of the game. 

GVGAI games are in general stochastic and can be played by agents via an API. In its original implementation, GVGAI provides a forward model that allows the agent to simulate future states given the current game state and an action to apply. This setting has been employed for the single and two-player tracks of the GVGAI competition~\cite{perez20162014}. The learning configuration, however, does not provide a forward model. In this scenario, agents are able to learn by repeatedly playing a given game, potentially becoming better at it as the number of episodes increases. The learning track uses the GVGAI\_GYM interface, implemented by Torrado et al.~\cite{Torrado_2018}, which integrates the GVGAI framework with the OpenAI Gym library for learning agents. Via this interface, the agent has access to the RGB observation of the current screen and the game score at each frame. The agent can then return an action to be executed in the game.

\subsection{Games} \label{ssec:games}

The GVGAI framework provides a large collection of games, with different difficulties. Similarly to the ALE framework~\cite{bellemare2013arcade}, GVGAI also has games with both sparse (score events are very rare) and dense rewards . The main difference is that GVGAI provides multiple levels for each game and it is easy to modify the rules and the rewards of the game and create new levels. Many games in GVGAI use stochastic elements, which make the learning problem harder than the games in ALE. The subset of games selected in this paper tries to capture different characteristics of GVGAI. We have not selected games with sparse rewards, as they require the usage of reward shaping or advanced exploration techniques, which was not the main goal of our experiments.

\begin{figure*}[t!]
  \centering
    \begin{subfigure}[b]{0.30\linewidth}
    \includegraphics[width=\linewidth]{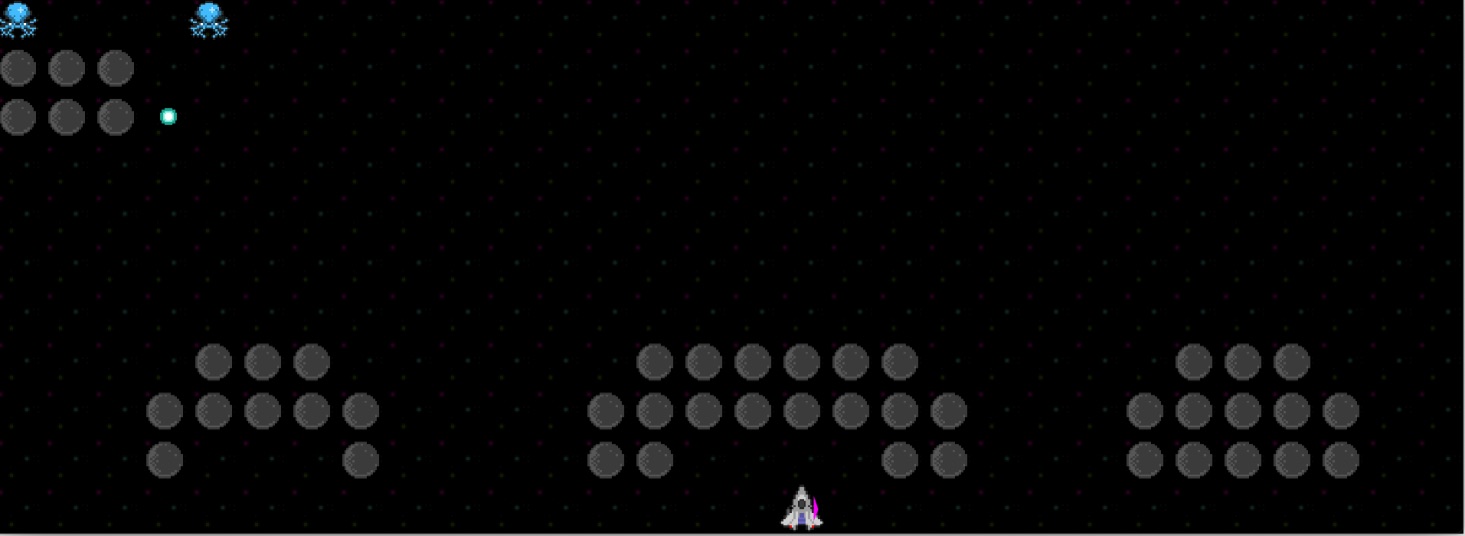}
    \caption{Aliens lvl0}
  \end{subfigure}
  \begin{subfigure}[b]{0.16\linewidth}
    \includegraphics[width=\linewidth]{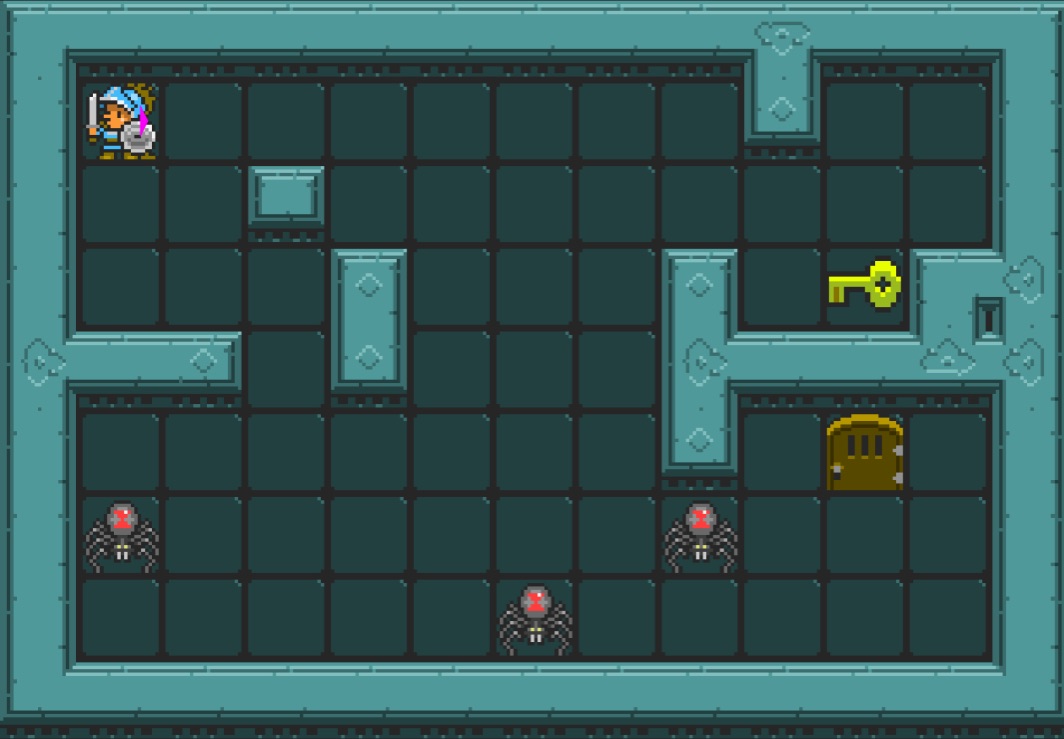}
    \caption{Zelda lvl0}
    \end{subfigure}
  \begin{subfigure}[b]{0.22\linewidth}
    \includegraphics[width=\linewidth]{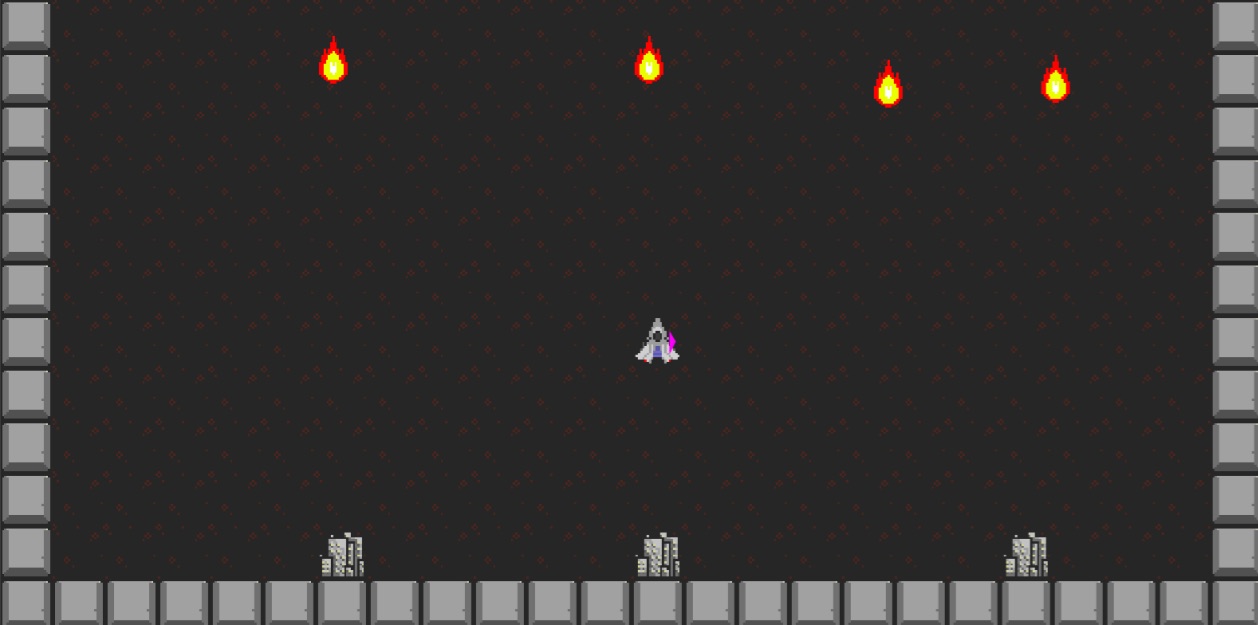}
    \caption{Missile Command lvl0}
  \end{subfigure}
    \begin{subfigure}[b]{0.26\linewidth}
    \includegraphics[width=\linewidth]{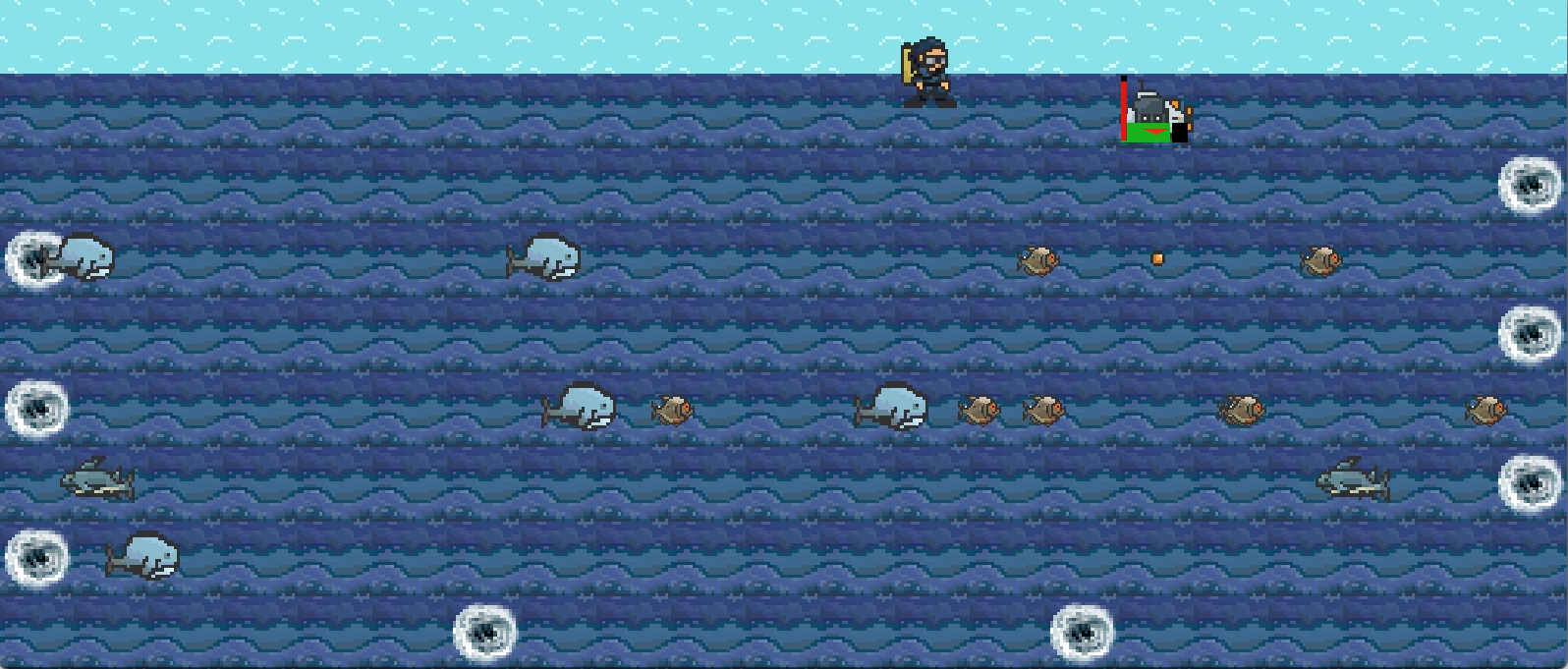}
    \caption{Seaquest lvl0}
  \end{subfigure}
  
      \begin{subfigure}[b]{0.30\linewidth}
    \includegraphics[width=\linewidth]{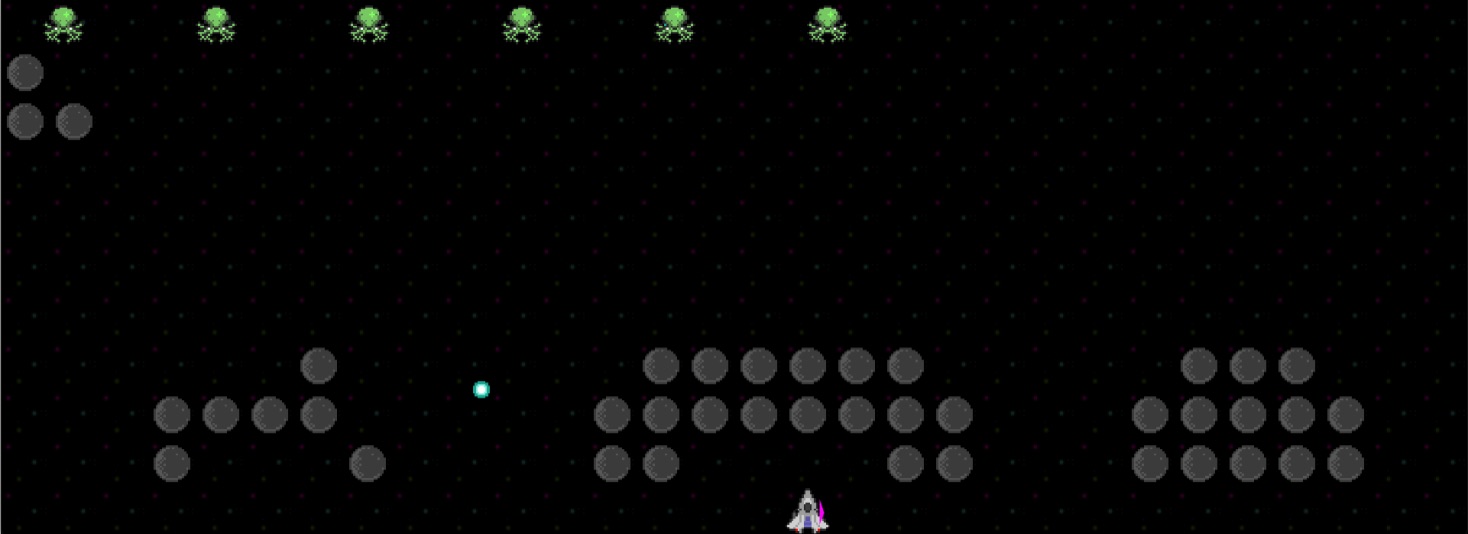}
    \caption{Aliens lvl1}
  \end{subfigure}
    \begin{subfigure}[b]{0.16\linewidth}
    \includegraphics[width=\linewidth]{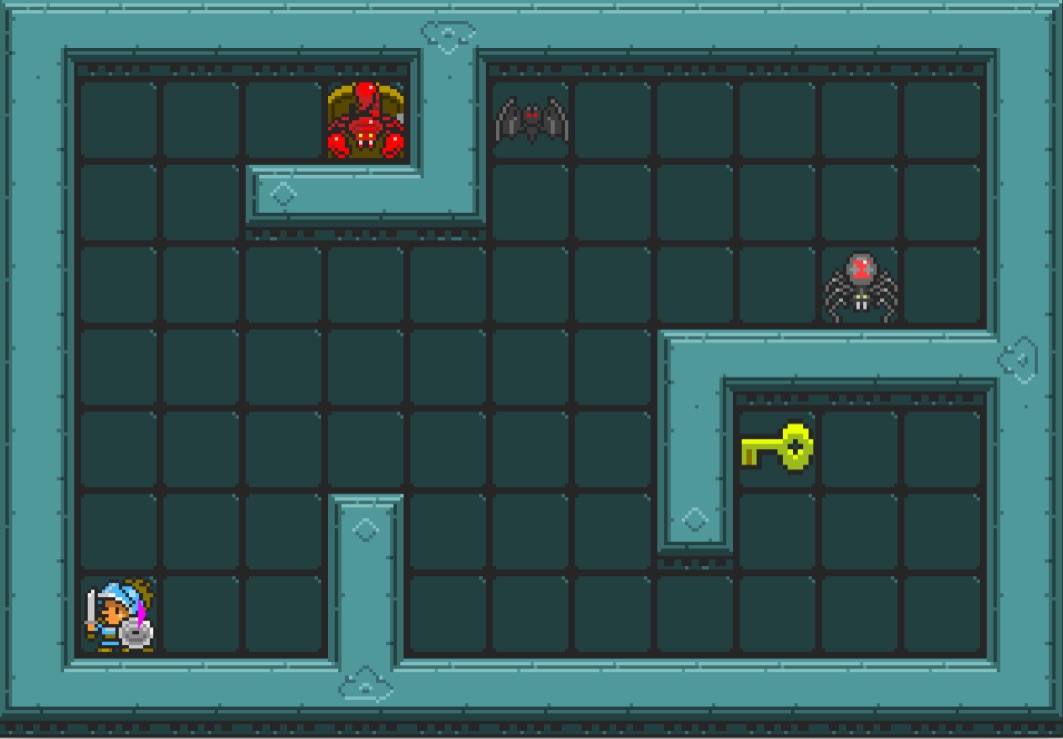}
    \caption{Zelda lvl1}
    \end{subfigure}
     \begin{subfigure}[b]{0.22\linewidth}
    \includegraphics[width=\linewidth]{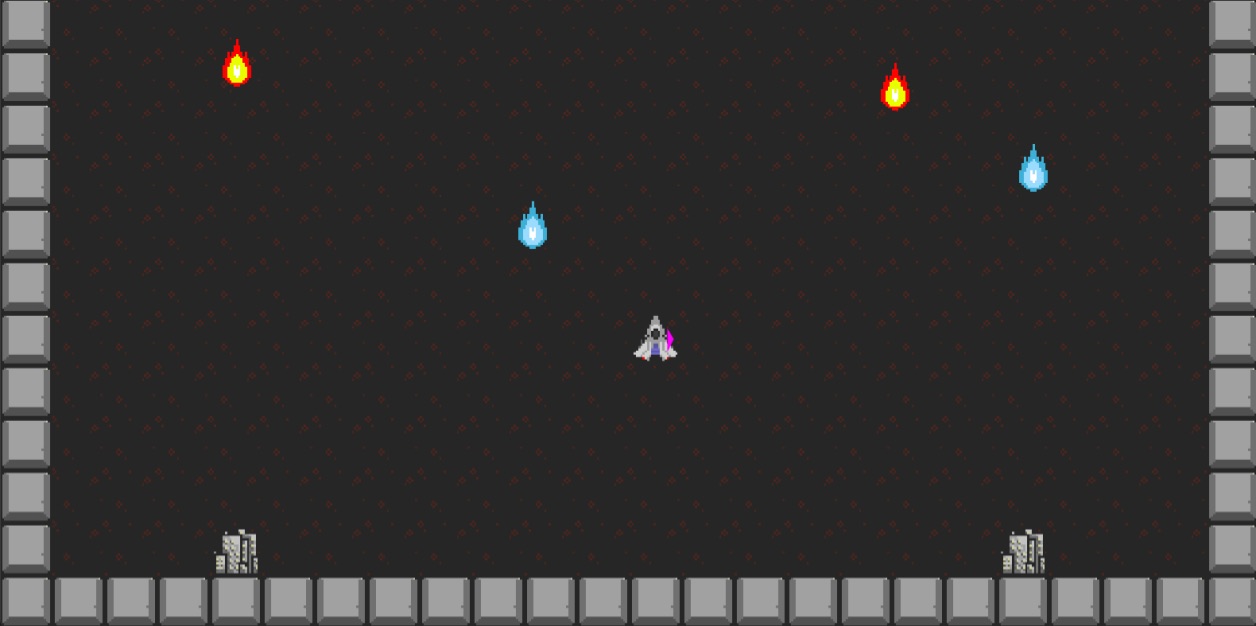}
    \caption{Missile Command lvl2}
  \end{subfigure}
    \begin{subfigure}[b]{0.26\linewidth}
    \includegraphics[width=\linewidth]{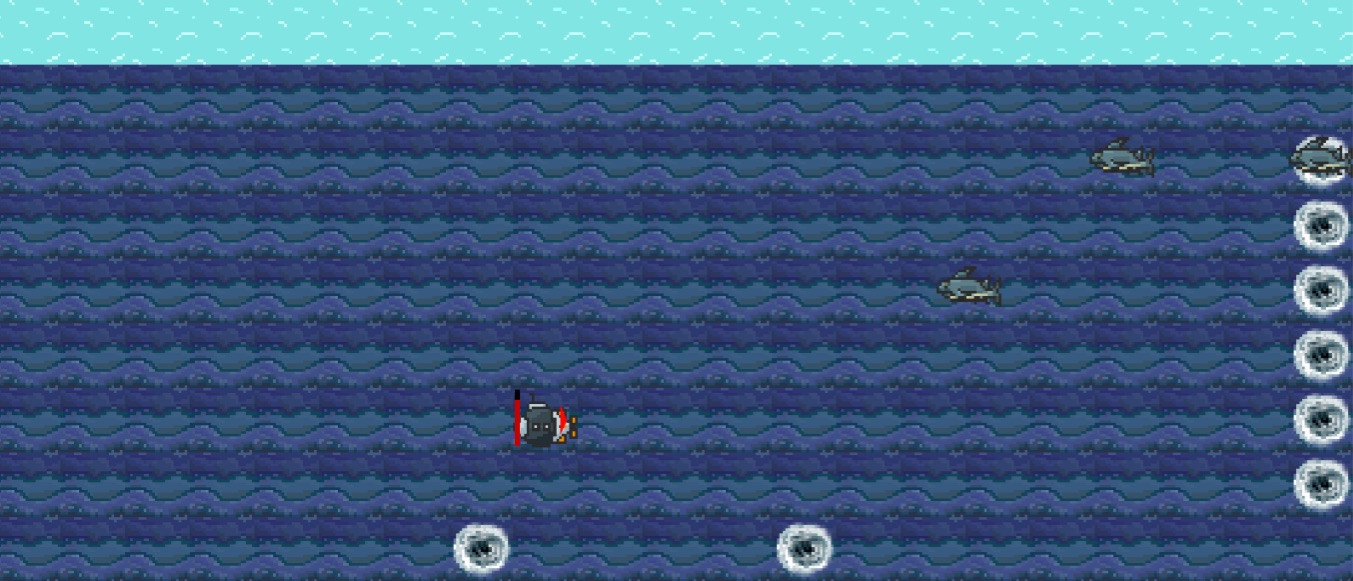}
    \caption{Seaquest lvl3}
  \end{subfigure}
  \caption{Example screenshots from the games used for this study. Aliens lvl0 has blue aliens, while lvl1 has slightly faster green aliens. Zelda lvl0 only has 3 enemies of the same type, while lvl1 has one of each type with different speed and action-repeat. Missile Command lvl0 only has red missiles, while lvl2 also has faster blue missiles. Seaquest lvl0 has a large selection of fish and 2 holes at the bottom from where the divers emerge, while lvl3 has less fish and the divers from the bottom holes spawn $5$ times more frequently.}
  \label{fig:games}
\end{figure*}

Figure~\ref{fig:games} shows the four selected games from the GVGAI framework, which were used in this study. All these games use $6$ actions (\texttt{Up}, \texttt{Down}, \texttt{Left}, \texttt{Right}, \texttt{Use} and \texttt{NIL}). All games have dense rewards, although their frequency and magnitude varies. The sprite graphics used by each game are also different, and so are the rules and the winning conditions. Thus, these games have been selected in order to provide a challenging scenario for the learning agents. 

An objective of this work is to analyze how the presence of features in the training set can influence the generalisation across different levels. Part of the experimentation work of this paper was performed only on the first 2 levels for each of the selected games. However, generally speaking, they do not capture all the existing game features, leading to poor generalisation. This subsection describes the different training sets employed and compared in our experiments, with the reasoning behind them.

The description of the selected games is as follows:

\paragraph{Aliens} This game is a version of Aliens (based on Atari's Space Invaders), where the player, rather than being locked into the bottom of the screen, can move in all directions. Touching the base kills the player instantly, which gives a reward of $-2$. The same happens when the player gets hit by the alien shots, with a reward of $-1$. Each alien gives $2$ points when killed and each base (by shooting) gives $1$ point when destroyed. As the starting location of the player is close to the bases, a random (or less intelligent) agent dies early in the game. This game was used in the 2018 and 2019 Single-player Learning Track competitions and can be found in the framework under the name of "cec1". The first 2 levels of this game capture all the difficulties that can come up in this game, as they include both types of aliens (set \texttt{aliens}). We refer to \texttt{aliens0} as the training set which only contains the first level for training.

\paragraph{Zelda} In order to win the game, the player has to collect a key ($1$ point) and then take it to the door (another point, if player has the key). There are uniform randomly moving non-player characters (NPC), that the player can kill ($2$ points each), but if they collide with the player, then the game ends and the player receives a reward of $-1$ points. The main difficulty for learning agents in Zelda comes from having $3$ different opponent types, all having different sprites with different speed and action-repeats (the same action gets executed for multiple frames). Zelda is the hardest game to train on. The first level only contains 3 enemies of the same type, while the second contains 3 enemies, but one of each type. When trained on these 2 levels together (set \texttt{zelda}), learning progress was poor, resulting in low scores and no wins. We investigated various setups and we found that when there are too many different types of enemies in the training set then none of the algorithms could learn good policies in $50$ million interactions. Set \texttt{zelda0} only uses the first level for training and \texttt{zelda2} uses the first level and the fourth, a level which contains the same enemy type as on the first one with an additional type. We found that training on \texttt{zelda2} was the only pair of levels in Zelda that resulted in completing the levels consistently. 

\paragraph{Missile Command}
The player controls a spaceship which has to stop incoming missiles before they reach and destroy the cities located at the bottom of the screen. Each one of the destroyed missiles is worth $2$ points, which can be eliminated by shooting at them (the player needs to be looking towards the missile and next to it). If a city is destroyed, $1$ point is lost. This game is deterministic, as the missiles follow the shortest path towards the cities. The game by default has $2$ types of missiles, which move with different speeds. The first $2$ levels contain only red missiles, while later levels have slightly faster blue missiles as well. The set \texttt{mc} contains the first 2 levels, which only have red missiles. To capture more features \texttt{mc2} has a level (lvl2) that also contains blue missiles along with the red missiles from the first level. \texttt{mc0} was trained on lvl3 only, which includes both types of missiles.

\paragraph{Seaquest}
The player controls a submarine that must rescue divers from the bottom of the sea and take them to the surface. Aquatic animals make the player lose the game when touched by them, and the submarine must watch its oxygen level, which decreases every certain number of time steps. If it gets depleted, the player loses the game, but it can be refilled if the player goes back to the surface. Animals can be eliminated by shooting at them, which gives the player $1$ point each. Taking $4$ divers at once to the surface gives $1000$ points (no points received if taken in smaller batches). This game is highly stochastic, as the spawn of animals and divers (including the movement of the latter) happens uniformly at random with different probabilities. Set \texttt{seaquest} uses the first 2 levels, which contain divers and various types of fish. \texttt{seaquest0} just uses the first level, while \texttt{seaquest2} uses level 3, which spawns divers more frequently. Level 3 facilitates getting high scores by spawning $5$ times more divers than other levels.

\section{Related Work} \label{sec:lit}

\subsection{Deep Reinforcement Learning}
Reinforcement Learning (RL) is a form of machine learning, where an agent interacts with its environment. The problem is typically framed as a Markov Decision Process, which allows us to study algorithms to solve such problems. The agent is presented with a state $s_t \in S$, where $S$ is the set of possible states at each step and has to select an action $a \in A$, where $A$ is the set of available actions and in response the environment returns the next state $s_{t+1}$ with a reward $r_{t+1} \in \mathbb{R}$. The objective of the agent is to reach the highest discounted cumulative reward in the environment.

Tabular methods represent values for exact state and action combinations. They require a table of size $|S| \times |A|$, which becomes problematic for most problems due to large action and state spaces. Neural Networks have shown great performance recently for function approximation, which can be used to approximate value functions and policies in RL. 

Deep Reinforcement Learning is the combination of Neural Networks with RL algorithms, which has shown superhuman performance on various benchmarks solely training from RGB pixel input. The first successful work using Deep RL was the Deep Q Network~\cite{mnih2015human}, which reached human level performance on the Atari Learning Environment (ALE) just using the RGB pixels as input. A Deep Q-Network (DQN) was running only a single environment at a time, which can be a bottleneck for modern computers, as the GPU is idle a lot of the time while waiting for the agent to collect a batch of experience. The Asynchronous Advantage Actor-Critic  (A3C)~\cite{mnih2016asynchronous} and it's synchronous version (A2C) have shown that Deep RL can be parallelized and similar or even better policies can be learned, while using less wall time. Actor-critic algorithms maintain $2$ networks, one is the actor executing the actions and the other is the critic, which approximates the value function and helps the actor in the learning process (by estimating a value function i.e: the advantage function to reduce the variance of the updates). In practice the actor and the critic share the same Neural Network body and they only get separated at the final layers. The A2C algorithm spawns multiple workers that operate in parallel in the environment and after a determined number of steps, make a synchronous update to the master network. After this, all workers can update their own parameters. A3C makes asynchronous updates, so the updates that the network gets are always from one agent's experience. However, A2C has shown to be more sample efficient than A3C, as all agents' experience are updated at the same time as a single batch, making the update more diverse. 

A3C has shown great results on the 2016 ViZDoom competitions where both $1^{st}$~\cite{Wu2017TrainingAF} and $3^{rd}$~\cite{ratcliffe17} place agents were based on A3C. Many researches uses the Advantage Actor-critic framework as their policy optimization algorithm. The option-critic~\cite{bacon2016optioncritic} uses a modified actor-critic algorithms to learn options (temporally abstracted actions). The UNREAL~\cite{jaderberg2016reinforcement} agent adds a replay buffer to replay rewarding situations more frequently and adds an auxiliary task to the agent, which forces it to learn how its actions influence its environment.
The A2C algorithm has been used in this work, which has a comparable performance to the state-of-the-art methods. Novel algorithms like PPO~\cite{schulman2017proximal} and IMPALA~\cite{espeholt2018impala} are generally more stable as they do not allow large changes in the policy in a single update, but for this reason they also need more training time. We chose the A2C algorithm to perform our experiments, due to its simplicity and good performance in previous works in both ALE and GVGAI.

\subsection{Generalisation over levels}
One of the  main objectives of this paper is to investigate the possibility of training an agent only on a maximum of 2 given levels, that transfers to new unseen levels without further training. 

 Most work on RL uses a single level for training, which also serves as the evaluation set. Farebrother et al.~\cite{farebrother2018generalization} used DQN to learn policies in ALE and evaluated how they generalize to modified version of the games. To improve generalisation they applied regularization techniques on DQN, which seemed to help. Some recent works have suggested the usage of procedurally generated levels to increase the training set~\cite{cobbe2019quantifying, justesen2018illuminating, cobbe2019leveraging}. Augmenting the set of training levels has shown to result in more robust policies, when trained using RL algorithms. Unfortunately augmenting the training levels is not always possible or feasible, for example in real world applications one might not have the possibility to create thousands of levels to learn simple tasks. This motivates our research to investigate the extent to which generalisation happens when trained on a very limited set of levels. We selected various scenarios to test ranging from training on a single level to training on a pair of levels, which best describes the games. Evaluation is done in zero-shot, without any further training on the test levels.

\subsection{Learning Agents in GVGAI}

The GVGAI competition is organised in various tracks, one of them being the Single-Player Learning track, which runs since 2017. Before the competition took place, Samothrakis et al.~\cite{Samothrakis15} showed the possibility of training learning agents in $10$ GVGAI games by means of Neuro-Evolution. Also Kunanusont et al.~\cite{kunanusont2017general} developed enhanced Deep Q-Learning agents that were able to play GVGAI games via screen capture, showing that the approach was able to learn to play both stochastic and deterministic games, increasing score and victory rate on them.

In 2018, Torrado et al. \cite{Torrado_2018} integrated the GVGAI framework with OpenAI's gym, in order to facilitate training with deep RL agents implemented in this library. In that work, the authors benchmarked two versions of DQN and an A2C agent, and trained them on the first level of a set of $8$ games without evaluating them on the other levels. Additionally, Justesen et al.~\cite{justesen2018procedural} also implemented A2C in a training environment integrated with procedural generation of levels. In this setting, levels of increasing difficulty were provided to the agent in response to agent's learning process.


\section{Methods} \label{sec:methods}
In this section we present the variants of A2C, their neural network architectures and the experimental setup used in this work.

\subsection{Agents}
Three versions of the A2C algorithm were used in our experiments, each with various training levels, which are described in Section~\ref{ssec:games}. 

\textbf{A2C} is an implementation of the synchronous version of the Advantage Actor-critic algorithm~\cite{mnih2016asynchronous}, which is used as the base of the other agents. The agent referred to as \textbf{GAP}, uses Global Average Pooling (GAP) over the last convolutional layer's activation, instead of a flattening layer. This way arbitrary input dimensions can be fed into the network, as the network does not depend on the width and the height of the input. The shortcoming of the method is, that it loses the spatial information about where the convolutional filters activate and works as a bag of features. The motivation behind using it, is that it has shown good results in preliminary experiments. It has also been used in winner's entry of the 2019 CEC GVGAI learning track competition. GAP makes it possible to work with various input dimensions (width and height of the input), which tackles the problem of having various input dimensions between levels or games (not the focus of the current work). The top entries and the baseline agents could not be evaluated in the earlier competitions~\cite{perez2019general}, due to the competition levels having different input sizes (width and height) between levels of the same game (this is not the case in our selection of games). GAP has regularization effects and have been used in image classification networks, we were hoping that it might learn different features, than other networks and might help in generalisation.

\textbf{PopArt} is a reward normalization technique introduced by van Hasselt et al.~\cite{van2016learning}. As many games have different reward distributions in scale and frequency the same hyperparameters might not perform well on all games. PopArt learns the mean $\mu$ and standard deviation $\sigma$ of the return and use them to normalise the targets. This alone makes the learning problem harder as these targets are non-stationary, so the critic's last weights and biases should be updated in a way that the output remain consistent after updating the statistic $\mu$ and $\sigma$.

\subsection{Network Architecture} 
The same Neural Network architecture has been used throughout the paper, with a small difference for GAP. The same network architecture has been used as by Mnih et al.~\cite{mnih2015human}, but with $2$ heads, one for the actor and one for the critic's output. We used $3$ convolutions layers with $32$ filters with kernel size $8$ and stride $4$ followed by 64 filters with kernel size $4$ and stride $2$ and finally $64$ filters with kernel size $3$ and stride $1$. In the case of A2C and PopArt, these get flattened and fed into a fully connected layer with $256$ units. In the case of GAP the width and height dimensions of the last activation gets averaged, which gives $64$ values, that gets fed into the fully connected layer without the need for flattening. The critic has a single output value, while the actor outputs a probability distribution over the action space (1 output for each action). The actions are sampled using a Normal Distribution over the action distribution.

For optimization we used the same loss function as used by Mnih et al.~\cite{mnih2016asynchronous} with the RMSProp optimization function. We used an exponential learning rate decay, every $1000$ optimization steps we multiplied the learning rate by $0.95$ with an initial learning rate of $7\text{e-}4$. For PopArt we used the same hyperparameters as were used by Hessel et al.~\cite{hessel2018multitask} $\beta =$ $3\text{e-}4$ and a lower bound of $1\text{e-}4$ and upper bound $1\text{e}6$.

\section{Experimental Setup} \label{sec:expsetup}

We used the games described in Section~\ref{ssec:games} for evaluating the methods described in the previous section. As mentioned above, the selected games have different reward distributions. In order to avoid providing domain knowledge about the games to the algorithm, we avoided reward clipping as it would change the objective of the agent, for example on Seaquest getting a $1000$ points for collecting 4 divers would not be as beneficial with clipping and the agent would learn a very different policy.

All the policies have been trained using $32$ workers on $16$ CPUs and a single GPU for the updates. The training was done for $50$ million frames per game on up to $2$ levels. Training levels were equally distributed between workers (thus, when training for $2$ levels on a single game, $16$ workers were used by level). The network takes in the RGB inputs from GVGAI\_GYM, with each pixel taking values in the range of $[0,255]$, which are normalized to $[0, 1]$. The environment frames are fed into the network without any further pre-processing (no downsampling or grayscaling) to avoid information loss over the different levels/games.

Reinforcement Learning agents tend to learn different policies each time they are trained due to having a random initialization for the weights, randomly sampling actions from their action distribution and random elements in the environment. Thus, we trained each algorithm on each game with $3$ different random seeds and averaged the results. For evaluation we used the final weight of each run and evaluated on all $5$ levels $20$ times to get a better estimate of the true final performance. We report the mean and the standard deviation of the final scores reached in the evaluations. As we run $3$ random seeds and $20$ evaluations, we aggregate the results over $60$ evaluation runs per agent per level. The results of the evaluations are shown in the tables below.

\section{Results} \label{sec:res}


All algorithms have been trained on the selected training sets $3$ times. It is important to mention that the stochasticity and large scales in rewards made a very small subset of these runs unable to converge to satisfactory learning. On the one hand we want to highlight the shortcomings of the tested methods with this type of environments. On the other hand, we wish to present plots that are representative of how learning takes place (when it does), and averaging non-convergent runs would make this not possible. Hence, in the affected experiments (GAP and A2C in \texttt{seaquest2} and PopArt and GAP in \texttt{zelda2}, only 2 runs were counted into the evaluations and 2 curves averaged in the figures. Training was done for $50$ million time steps over $3$ random seeds per algorithm. The plots only show the first $20$ million frames for clarity, the remaining $30$ million frames did not show improvements for any of the games. Note that the scale of the rewards and lengths are not comparable between training sets as their scales differ across levels. The values in the tables show the mean episode scores and win rates and the values in parentheses show the standard deviation, the shaded areas indicate the levels used for training for that row. During training, statistics of the runs were collected after every $200$ optimization steps (every $32,000$ environment steps) and each point in the plots show the running mean over the last $100$ episodes.

\begin{figure}[!t]
	\begin{center}
	\includegraphics[width = .24\textwidth]{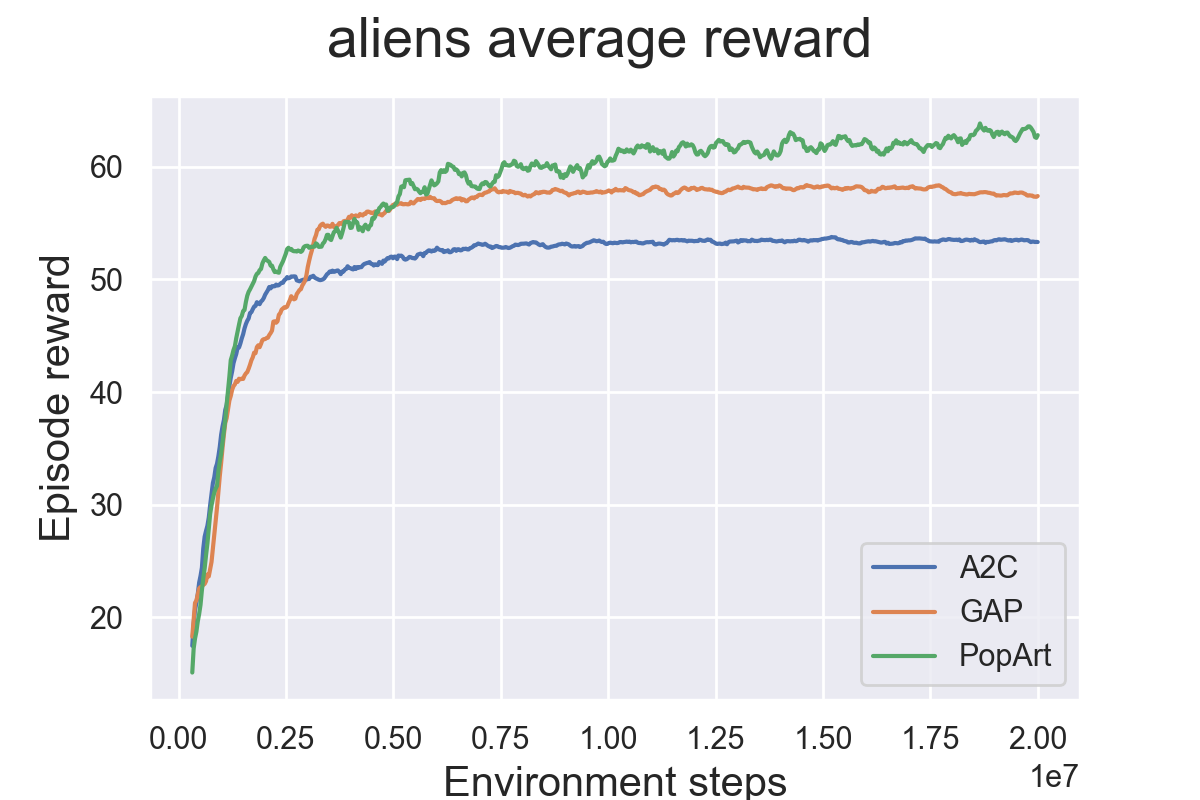}
	\includegraphics[width = .24\textwidth]{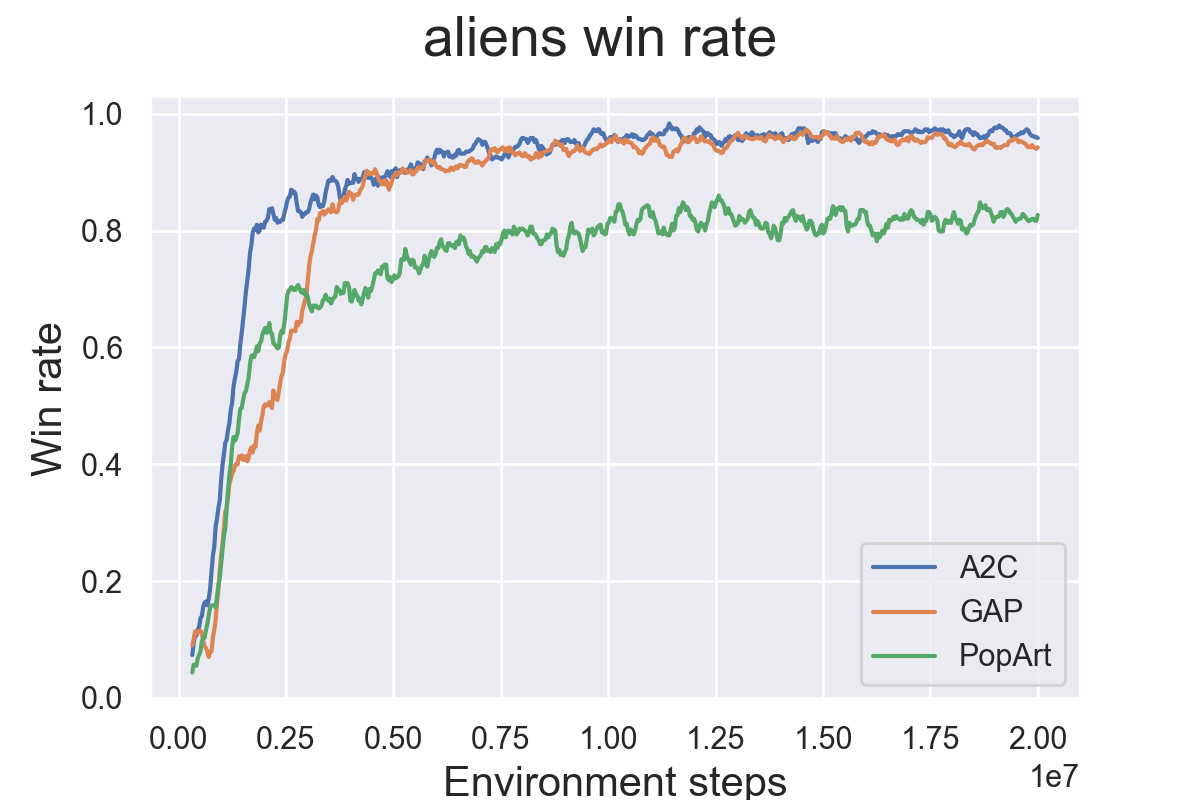}
	\caption{Training results for Aliens. The plots show the running mean of the episode rewards and the win rate, during training on \texttt{aliens} (lvl0 and lvl1).}
	\label{fig:aliens}
	\end{center}
\end{figure}

The results of training on Aliens are shown in Figure~\ref{fig:aliens} and Table~\ref{tab:aliens}. The figures show that PopArt is slightly getting higher scores, while it gets the lowest win rates. The evaluations in Table~\ref{tab:aliens} show that, when the agent was trained on only the first level it could do well on the second level, even thought the aliens moved slightly faster and used a different sprite (green instead of blue). The unseen levels provided some challenge in general, but the trained policies occasionally managed to win and get reasonable scores. The evaluation levels only had slight variations compared to the first 2 levels, like having the bases in different positions or no bases at all. GAP got the highest win rate over all levels by a small margin over A2C. 

\begin{figure*}[!t]
	\begin{center}
	\includegraphics[width = .24\textwidth]{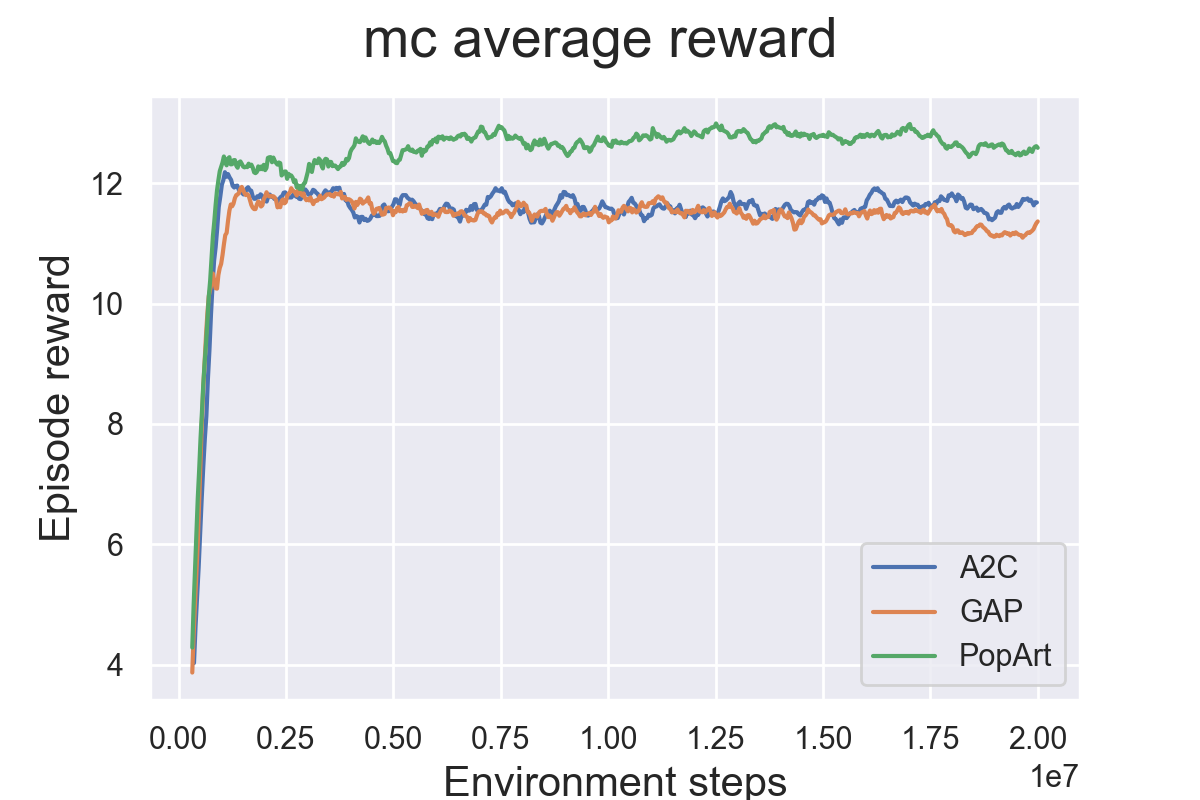}
	    \includegraphics[width =
	.24\textwidth]{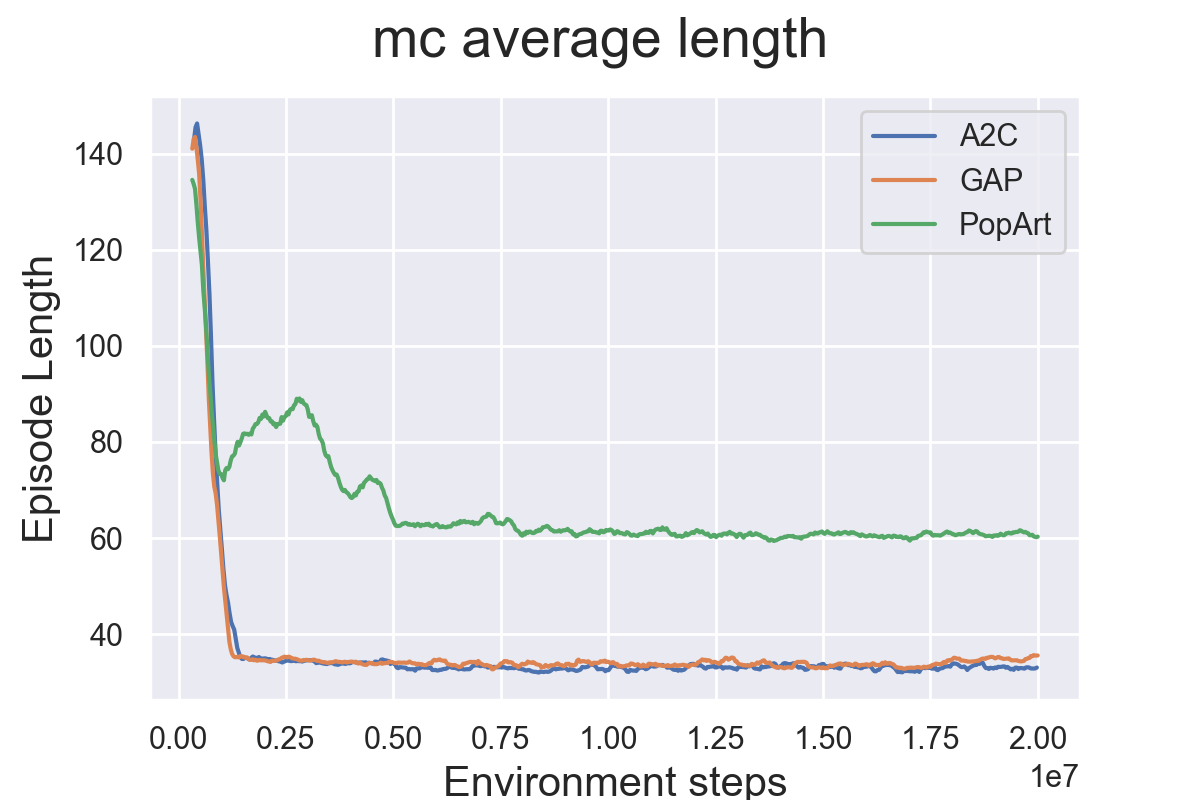}
	\includegraphics[width =
	.24\textwidth]{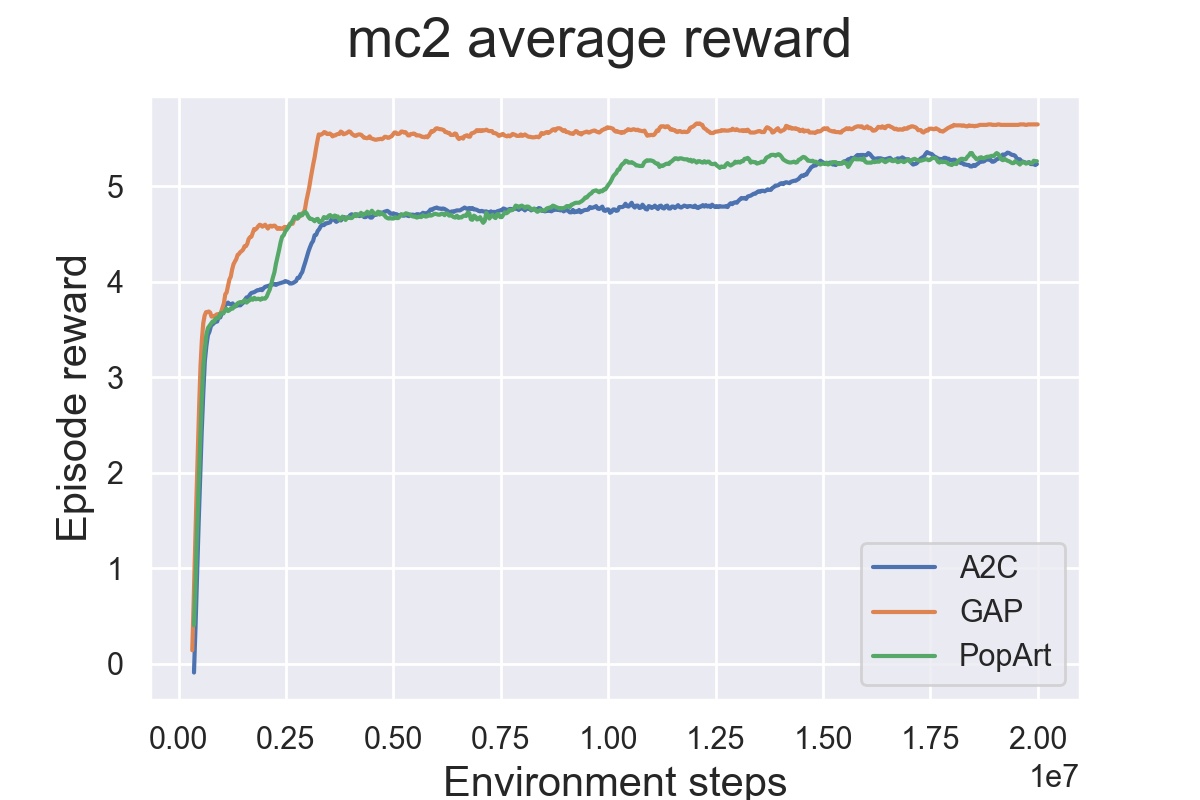}
		\includegraphics[width =
	.24\textwidth]{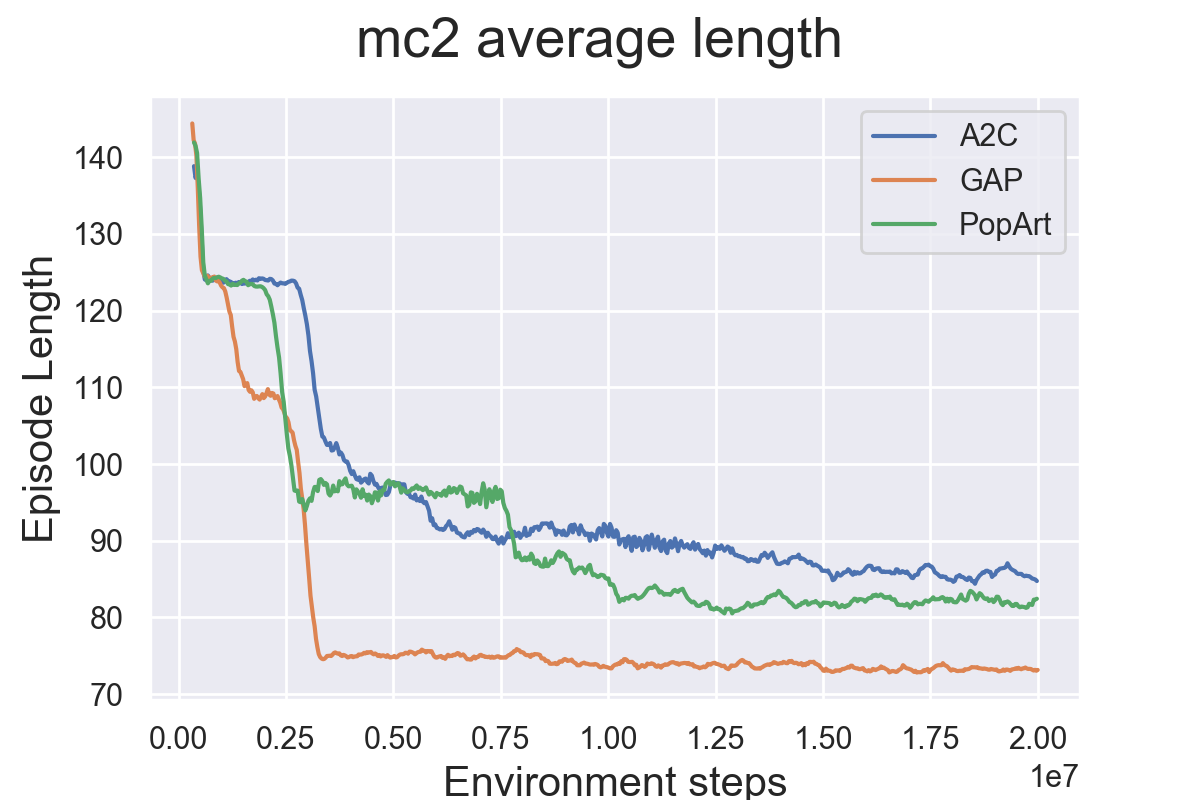}
	\caption{Training on Missile Command. The two plots on the left show results for training set \texttt{mc}, right plots for \texttt{mc2}. Plots show running average of the episode rewards and length. Note that the scale of the rewards differ between the training sets.
	}
	\label{fig:missilecommand}
	\end{center}
\end{figure*}

Missile Command seems to be the easiest game for training, that was used in this experiment. During training all agents reached a $100\%$ win rate after a few million steps. Figure~\ref{fig:missilecommand} shows that on \texttt{mc} PopArt found a slightly better policy, which results in a slightly higher average reward, but takes longer. On \texttt{mc2}, the agents performed rather similarly, GAP got higher rewards slightly faster. Table~\ref{tab:missilecommand} shows that almost all agents got maximum scores for evaluation on the levels they were trained on. As \texttt{mc} is deterministic the learned policies did not perform well on the evaluation levels. The last level lvl4 is misleading as there are more cities than missiles, so the player always wins. Comparing the 3 training sets, it seems that having 2 levels helped the agents reach higher scores on the evaluation levels. Having both red and blue missiles does not seem to improve generalisation, but it may be due to the deterministic events in the game.

\begin{figure*}[!t]
	\begin{center}
	\includegraphics[width = .24\textwidth]{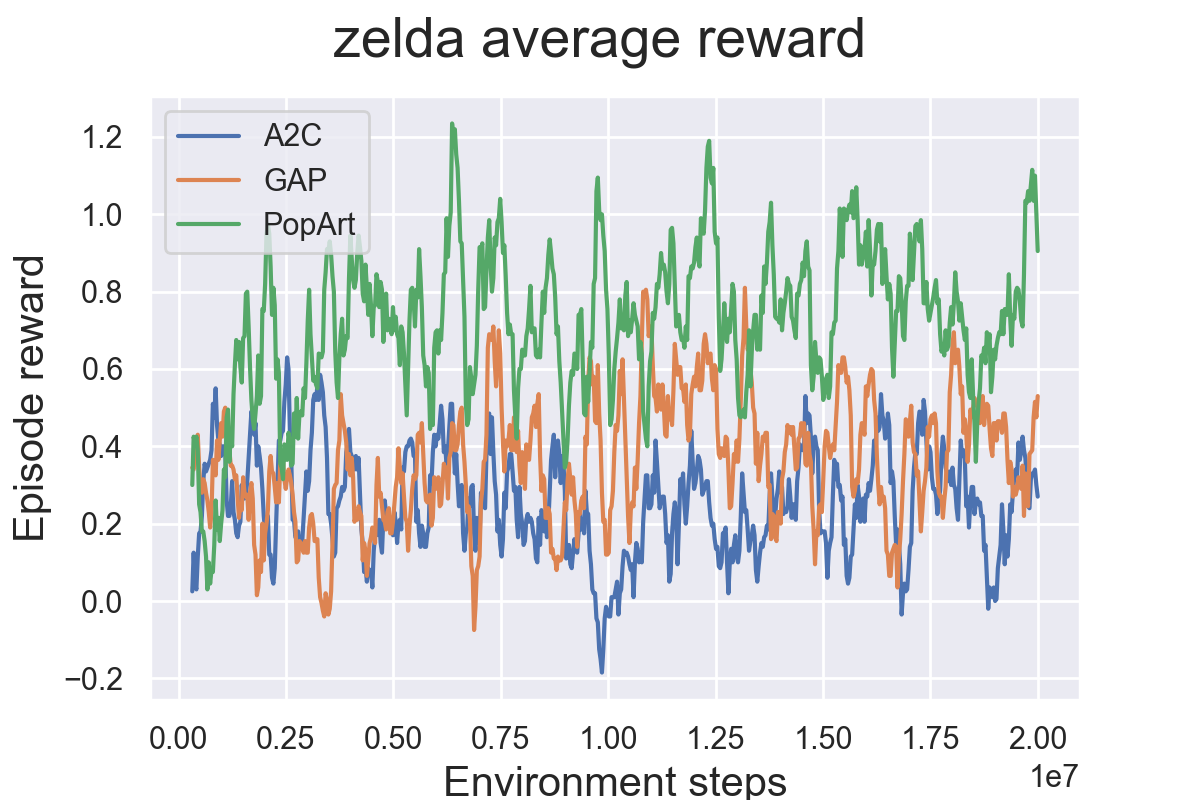}
	\includegraphics[width =
	.24\textwidth]{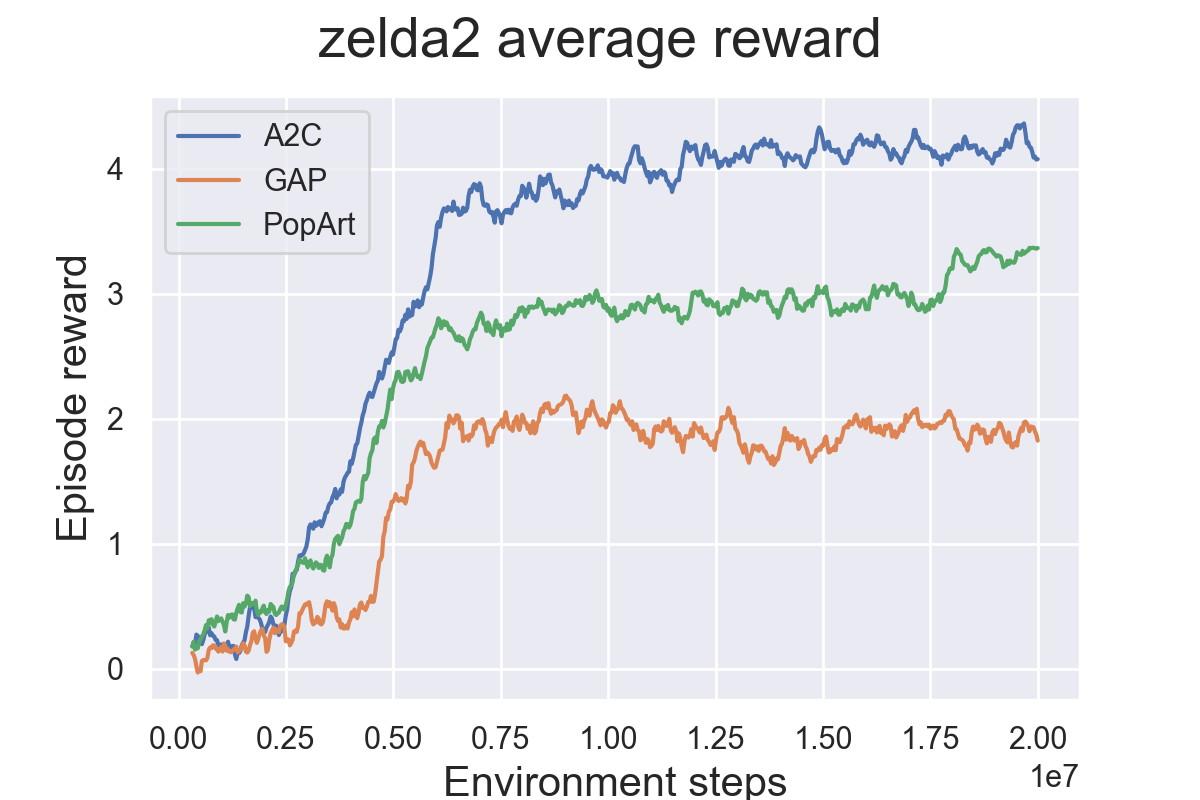}
		\includegraphics[width =
	.24\textwidth]{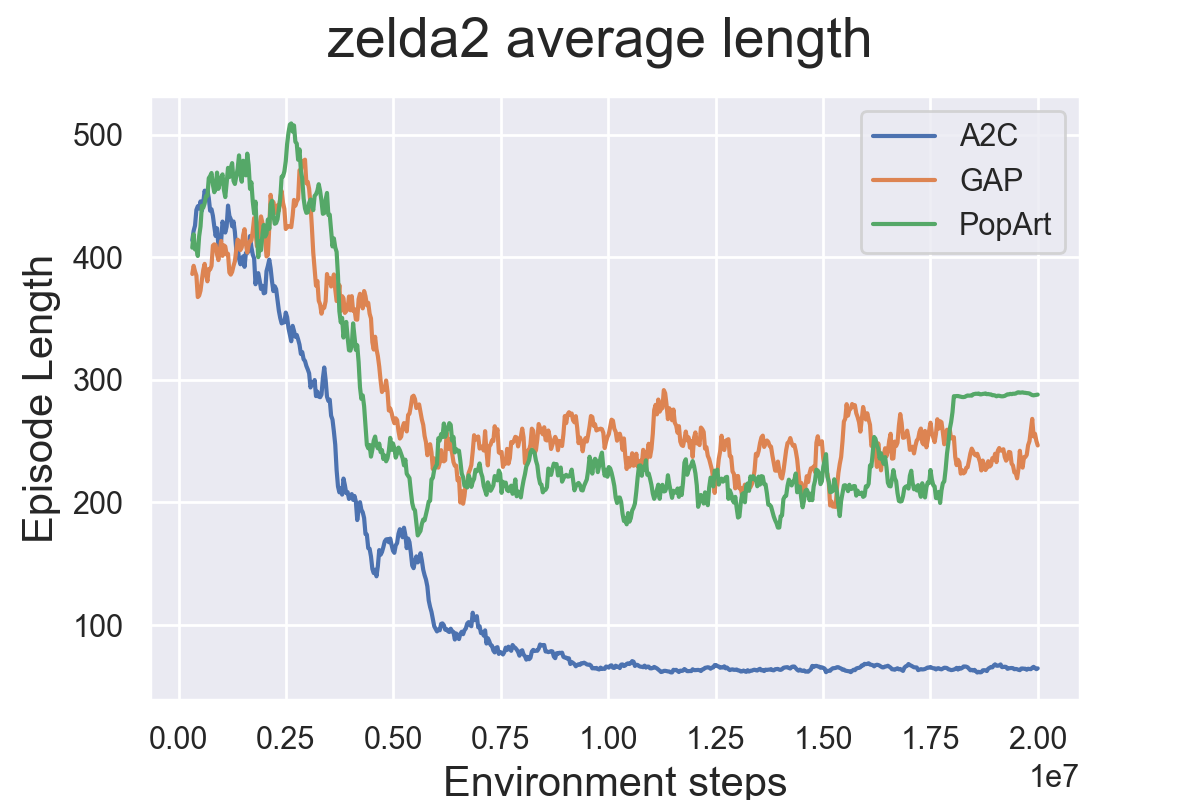}
		\includegraphics[width =
	.24\textwidth]{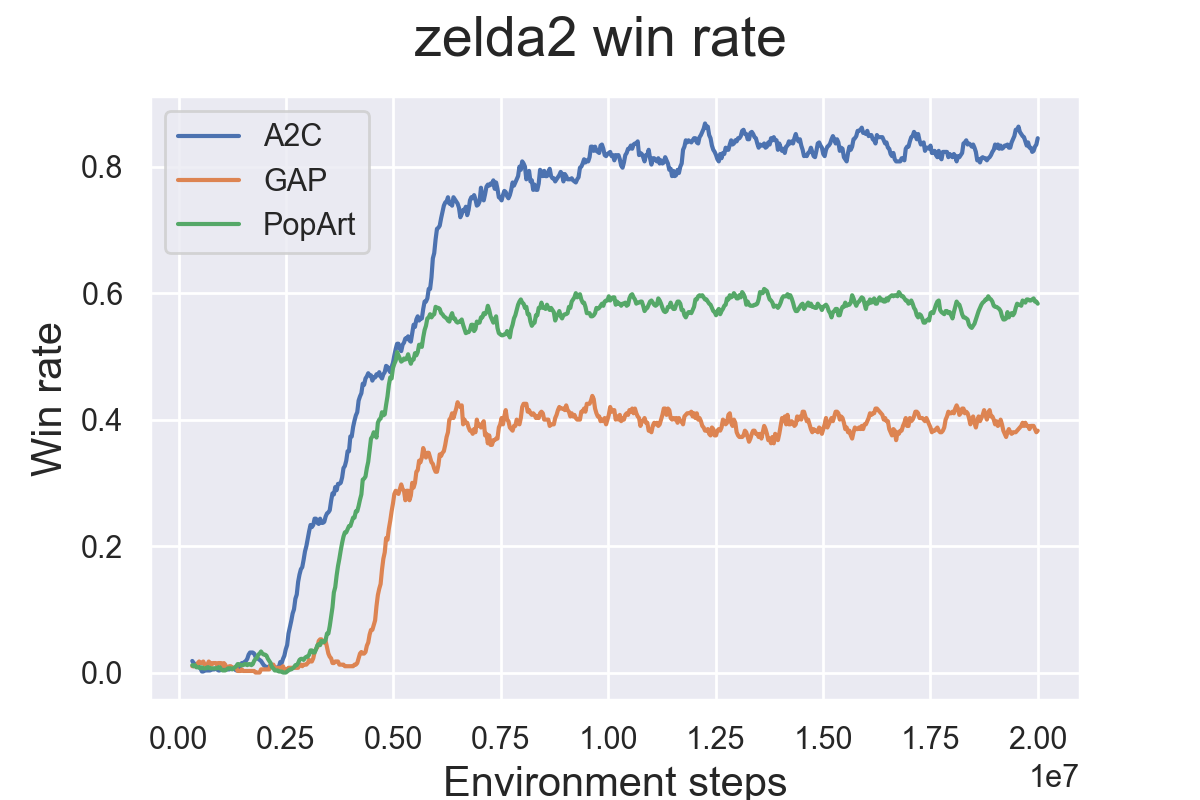}
	\caption{Training results on Zelda. The first plot shows the running mean of the episode reward using \texttt{zelda}. The other plots show the training performance using \texttt{zelda2}.}
	\label{fig:zelda}
	\end{center}
\end{figure*}

One of the problems we discovered when training on Zelda is that, due to having multiple opponents with different movement patterns, training became extremely hard. The first plot on Figure~\ref{fig:zelda} shows that when we trained on the first 2 levels (first level only contains 1 enemy type, while the second contains all 3) none of the algorithms could learn a policy that would result in consistent scoring. The plots for \texttt{zelda2} on Figure~\ref{fig:zelda} show that training on lvl0 and lvl4 resulted in a much better policy in terms of scores and win rates. Interestingly, A2C outperformed PopArt during training and GAP performed the worst. Table~\ref{tab:zelda} shows that agents failed to win in levels they were not trained on. Even on the trained levels the agent could not achieve a $100\%$ win rate during evaluation. The best agent was PopArt on the training set \texttt{zelda2}, slightly getting more wins ($34\%$) in this case. We found lvl1 causing difficulties to train on, when it was used in the training set, none of the agents could learn a reasonable policy. PopArt reached the highest score on average when it was trained on lvl1, but it never managed to win the level.

\begin{figure*}[!t]
	\begin{center}
		\includegraphics[width =
	.24\textwidth]{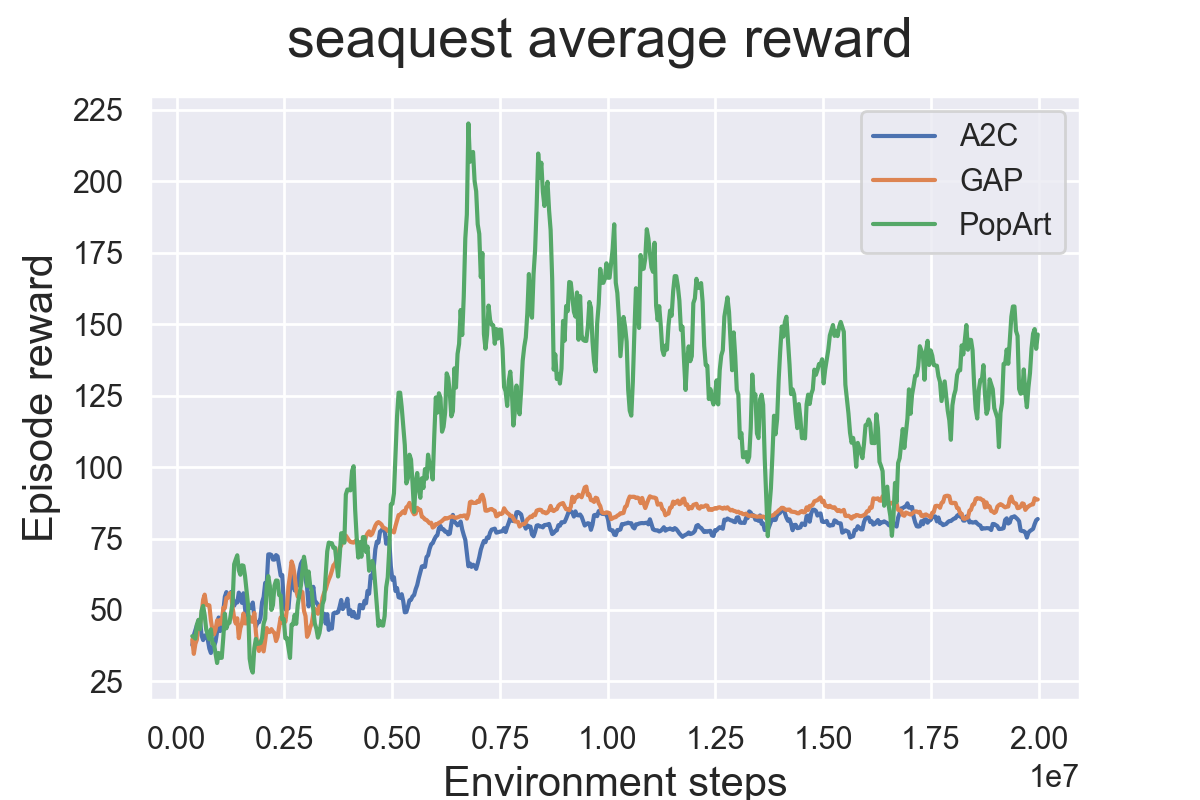}
		\includegraphics[width =
	.24\textwidth]{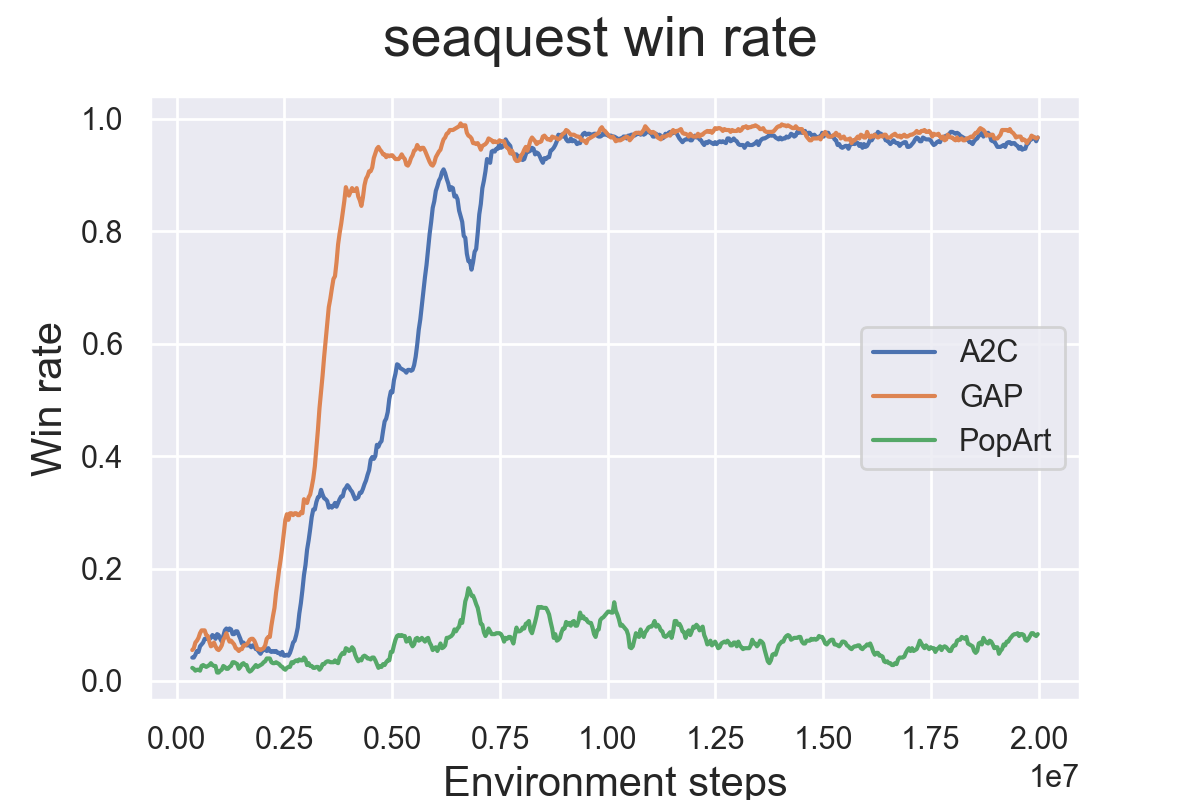} 
		\includegraphics[width =
	.24\textwidth]{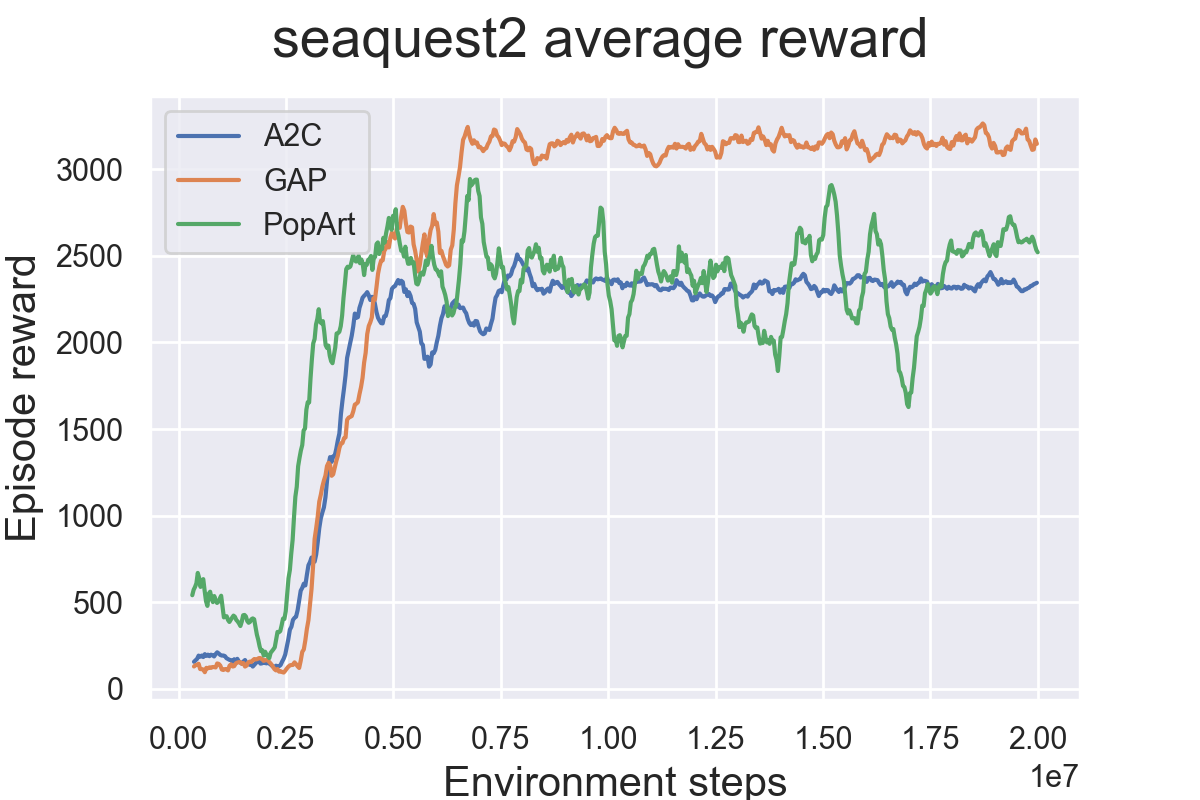}
		\includegraphics[width =
	.24\textwidth]{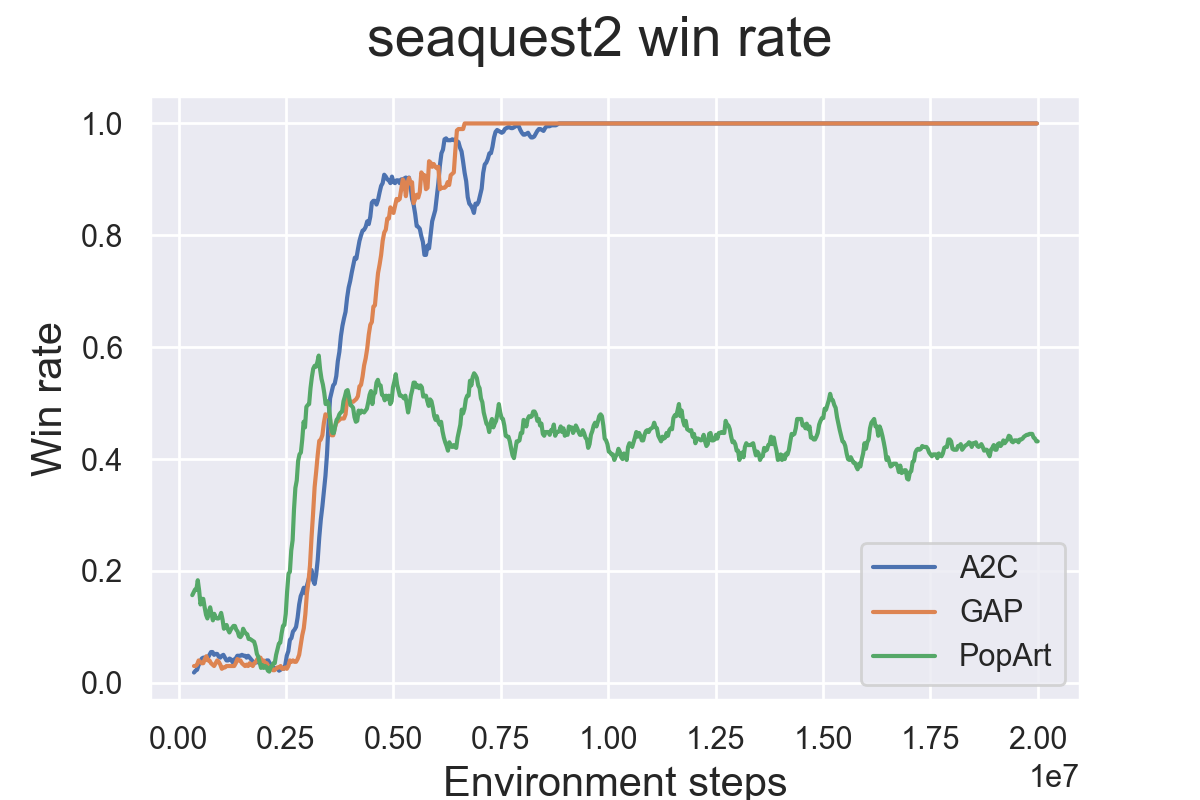}

	\caption{The two plots on the left show running average reward and win rate during training on \texttt{seaquest}, right plots for \texttt{seaquest2}. A2C and GAP only average 2 runs on the latter, as they both failed once to learn good policies.}
	\label{fig:seaquest}
	\end{center}
\end{figure*}

Finally, the results of Seaquest can be seen in Figure~ \ref{fig:seaquest} and in Table~\ref{tab:seaquest}. Seaquest presents the challenge of exploration and exploitation, as a local optima is to kill fish in the water, which gives $1$ point each, or take the risk and collect $4$ divers before running out of oxygen, which is worth a $1000$ points. When agents were trained on the first $2$ levels, they rarely managed to collect $4$ divers with a single tank of oxygen. As rewards were not clipped, A2C and GAP got quite unstable in some runs, both resulting in one training run (as mentioned above) where learning did not take place. PopArt was more stable, as it normalized the reward, which made it less sensitive to large changes in the scale and the frequency of the rewards. PopArt clearly outperformed the other algorithms in terms of scores, but not in win rates. A2C and GAP seem to be more cautious, they do not take as much risk as PopArt. To win the game the agent just has to avoid drowning or collision with the fish. When training on lvl0 and lvl1 the agent rarely figures out that it should collect 4 divers at the same time to maximise its score.
The evaluation results in Table~\ref{tab:seaquest} show, that having a level in the training set (lvl3) where the agent can more easily collect divers to earn points, helps in transferring that knowledge to the other levels. This confirms our hypothesis, that the quality of the training set matters and they influence the behaviour of the trained agent.

Our experiments show that training on a single level is easier to achieve high scores, but it is more likely to overfit, while having 2 levels improves the generalisation, but do not seem to provide enough experience to learn policies that could work well on unseen levels. Stochasticity definitely helps, as seen by the difference between Aliens and Missile Command, where the later is fully deterministic, resulting in the agents only memorising a trajectory and failing to win on the evaluation levels. Another interesting property that we observed is that getting a high score is not equal to winning the game. PopArt had the highest score in Seaquest and Aliens, but at the same time had the lowest win rate. PopArt's objective is slightly changed due to the learned normalization, which might cause it to care more about positive rewards than the end of the episode or a small negative reward. In Seaquest the agent does not get any reward for winning or losing, which might be the reason why PopArt only cares about maximizing score. It is not surprising that PopArt did not achieve the highest scores in every game, as even in van Hasselt et al.~\cite{van2016learning} it underperformed the baseline agent in various games, but overall it achieved a slightly higher median score over all Atari games. 
GAP is the most inconsistent algorithm, having a much larger variance in the evaluations and even during training (failed a run in both \texttt{zelda2} and \texttt{seaquest2} training sets). The inconsistency is likely to come from the fact that GAP loses the spatial information about which area of the state a convolutional filter activates. Surprisingly it can still outperform the other algorithms in a few cases, maybe due to the regularization capabilities acting as a bag of features and also having fewer weights (having less weights in the first fully connected layer, due to not using flattening).

\begin{table*}
\centering
\begin{tabular}{|ll|lllllllllll|}
\toprule
    Training set & Policy & \multicolumn{5}{l}{Score}
     & \multicolumn{5}{l}{Win rate} & avg.\\
 &  &  lvl0 &  lvl1 &  lvl2 &  lvl3 &  lvl4 &  lvl0 &  lvl1 &  lvl2 &  lvl3 &  lvl4 & win \\
\midrule
aliens0 & A2C & \cellcolor{gray!25}54.00(0.00) &  50.70(7.94) & -0.85(1.57) & -1.00(0.00) &  \textbf{19.55}(16.61) & \cellcolor{gray!25}1.00 &  0.70 &  0.00 &  0.00 &  0.25 & 0.39 \\
(lvl0) & GAP & \cellcolor{gray!25}61.10(8.26) & 48.20(15.54) & 1.90(12.04) & \textbf{32.95}(11.74) & -1.75(0.79) & \cellcolor{gray!25}0.95 &   0.35 &  0.05 &  0.65 &  0.00 & 0.40\\
 & PopArt & \cellcolor{gray!25}\textbf{63.25}(15.56) &  41.35(15.23) & -1.65(0.88) &  5.25(11.27) &  2.00(3.21) & \cellcolor{gray!25}0.90 &  0.15 &  0.00 &  0.05 &  0.00 & 0.22 \\ 
\midrule
aliens & A2C & \cellcolor{gray!25}53.95(1.74) &  \cellcolor{gray!25}53.15(4.11) & \textbf{13.37}(21.18) & -0.38(2.57) &  5.07(10.92) & \cellcolor{gray!25}0.97 & \cellcolor{gray!25}0.97 &  0.22 &  0.00 &  0.03 & 0.44\\
(lvl0, lvl1)     & GAP &  \cellcolor{gray!25}58.93(5.63) &  \cellcolor{gray!25}58.84(4.67) &  6.42(19.03) &  13.07(17.84) & -1.00(1.33) & \cellcolor{gray!25}0.94 &  \cellcolor{gray!25}0.96 &  0.14 &  0.26 &  0.00 & 0.46\\
 & PopArt & \cellcolor{gray!25}61.62(15.11) &  \cellcolor{gray!25}\textbf{61.17}(9.74) & -0.05(7.48) &  10.28(17.83) & -0.42(2.49) & \cellcolor{gray!25}0.77 &  \cellcolor{gray!25}0.80 &  0.02 &  0.25 &  0.00 & 0.37\\
\bottomrule
\end{tabular}
\caption{Evaluation results for \textit{Aliens}.}
\label{tab:aliens}
\end{table*}

\begin{table*}
\centering
\begin{tabular}{|ll|lllllllllll|}
\toprule
    Training set & Policy & \multicolumn{5}{l}{Score}
     & \multicolumn{5}{l}{Win rate} & avg. \\
 &  &  lvl0 &  lvl1 &  lvl2 &  lvl3 &  lvl4 &  lvl0 &  lvl1 &  lvl2 &  lvl3 &  lvl4 & win\\
\midrule
mc0 & A2C & -2.90(0.45) & -0.50(0.89) & -3.00(0.00) &  \cellcolor{gray!25}7.85(0.67) & -7.00(0.00) &  0.00 &  0.00 &  0.00 &  \cellcolor{gray!25}1.00 &  1.00 & 0.40  \\
(lvl3) & GAP & 0.20(2.19) & -1.00(0.00) & -3.00(0.00) & \cellcolor{gray!25}\textbf{8.00}(0.00) & -7.00(0.00) & 0.45 &   0.00 &  0.00 &  \cellcolor{gray!25}1.00 &  1.00 & 0.49\\
 & PopArt & -1.60(1.57) & -1.00(0.00) & -3.00(0.00) &  \cellcolor{gray!25}\textbf{8.00}(0.00) & -7.00(0.00) &  0.10 &  0.00 &  0.00 &  \cellcolor{gray!25}1.00 &  1.00 & 0.42  \\
\midrule
mc & A2C &  \cellcolor{gray!25}\textbf{8.00}(0.00) &  \cellcolor{gray!25}\textbf{16.00}(0.00) & -3.00(0.00) & -0.38(1.69) & -0.20(3.49) & \cellcolor{gray!25}1.00 &  \cellcolor{gray!25}1.00 &  0.00 &  0.02 &  1.00 & 0.60 \\
(lvl0, lvl1)     & GAP &  \cellcolor{gray!25}6.45(2.43) &  \cellcolor{gray!25}15.89(0.74) & -3.00(0.00) & -0.88(1.68) &  0.92(4.00) &  \cellcolor{gray!25}0.98 &  \cellcolor{gray!25}0.98 &  0.00 &  0.02 &  1.00 & 0.60 \\
 & PopArt &  \cellcolor{gray!25}7.45(1.17) &  \cellcolor{gray!25}\textbf{16.00}(0.00) & -3.00(0.00) & -1.22(1.01) &  \textbf{2.52}(4.04) &  \cellcolor{gray!25}1.00 &  \cellcolor{gray!25}1.00 &  0.00 &  0.00 &  1.00 & 0.60\\
\midrule
mc2 & A2C &  \cellcolor{gray!25}5.90(1.39) &  1.13(2.30) &  \cellcolor{gray!25}\textbf{4.95}(0.39) & -1.50(0.95) & -1.60(3.57) &  \cellcolor{gray!25}1.00 &  0.00 &  \cellcolor{gray!25}1.00 &  0.00 &  1.00 & 0.60  \\
(lvl0, lvl2)     & GAP &  \cellcolor{gray!25}6.00(1.43) & -0.90(0.77) &  \cellcolor{gray!25}\textbf{4.90}(0.54) & -1.87(0.50) & -4.00(3.79) &  \cellcolor{gray!25}1.00 &  0.00 &  \cellcolor{gray!25}1.00 &  0.00 &  1.00 & 0.60 \\
 & PopArt &  \cellcolor{gray!25}5.95(1.51) &  3.17(2.61) &  \cellcolor{gray!25}\textbf{4.85}(0.66) & -0.57(2.13) & -1.95(3.89) &  \cellcolor{gray!25}1.00 &  0.00 &  \cellcolor{gray!25}1.00 &  0.07 &  1.00 & 0.61   \\

\bottomrule
\end{tabular}
\caption{Evaluation results for \textit{Missile Command}.}
\label{tab:missilecommand}
\end{table*}

\begin{table*}
\centering
\begin{tabular}{|ll|lllllllllll|}
\toprule
    Training set & Policy & \multicolumn{5}{l}{Score}
     & \multicolumn{5}{l}{Win rate} & avg. \\
 &  &  lvl0 &  lvl1 &  lvl2 &  lvl3 &  lvl4 &  lvl0 &  lvl1 &  lvl2 &  lvl3 &  lvl4 &  win \\
\midrule
zelda0 & A2C & -1.00(0.00) & -1.00(0.00) & -1.00(0.00) & -1.00(0.00) &  \cellcolor{gray!25}4.00(2.68)  &  0.00 &  0.00 &  0.00 &  0.00 &  \cellcolor{gray!25}0.65 & 0.13\\
(lvl4) & GAP & -0.90(0.31) & -1.00(0.00) & -0.80(0.41) & -1.00(0.00) & \cellcolor{gray!25}\textbf{7.60}(1.23) & 0.00 & 0.00 &  0.00 &  0.00 &\cellcolor{gray!25}0.90 & 0.18\\
& PopArt & -0.35(0.67) & -1.00(0.00) & -0.90(0.31) & -1.00(0.00) &  \cellcolor{gray!25}5.50(1.43) &  0.00 &  0.00 &  0.00 &  0.00 &  \cellcolor{gray!25}0.95 & 0.19 \\
\midrule
zelda & A2C & \cellcolor{gray!25}-0.20(1.20) &  \cellcolor{gray!25}0.35(1.84) & -0.10(1.92) & -0.40(1.14) & -0.50(1.10)&  \cellcolor{gray!25}0.00 &  \cellcolor{gray!25}0.00 &  0.00 &  0.00 &  0.00 & 0.00  \\
(lvl0, lvl1)       & GAP &  \cellcolor{gray!25}0.60(1.96) &  \cellcolor{gray!25}0.10(1.74) &  0.00(1.86) & -0.30(1.17) &  0.40(2.09) &  \cellcolor{gray!25}0.05 &  \cellcolor{gray!25}0.00 &  0.00 &  0.00 &  0.00 & 0.01 \\
& PopArt &  \cellcolor{gray!25}0.30(2.25) &  \cellcolor{gray!25}\textbf{1.25}(2.47) &  \textbf{0.80}(2.17) &  \textbf{1.75}(3.55) &  0.50(1.85) &  \cellcolor{gray!25}0.00 &  \cellcolor{gray!25}0.00 &  0.00 &  0.00 &  0.00 & 0.00\\
\midrule
zelda2 & A2C &  \cellcolor{gray!25}3.37(2.31) & -0.43(1.31) & -0.95(0.29) & -1.00(0.00) &  \cellcolor{gray!25}4.10(2.03) &  \cellcolor{gray!25}0.72 &  0.00 &  0.00 &  0.00 &  \cellcolor{gray!25}0.77 & 0.30 \\
(lvl0, lvl4)       & GAP &  \cellcolor{gray!25}1.90(1.80) &  0.50(1.15) & -0.75(0.72) & -0.90(0.45) &  \cellcolor{gray!25}3.65(1.93) &  \cellcolor{gray!25}0.60 &  0.00 &  0.00 &  0.00 &  \cellcolor{gray!25}0.70 & 0.26   \\
& PopArt &  \cellcolor{gray!25}\textbf{4.05}(2.24) & -0.80(0.61) & -0.98(0.16) & -0.95(0.32) &  \cellcolor{gray!25}4.25(1.58) &  \cellcolor{gray!25}0.82 &  0.00 &  0.00 &  0.00 &  \cellcolor{gray!25}0.90 & 0.34\\
\bottomrule
\end{tabular}
\caption{Evaluation results for \textit{Zelda}}
\label{tab:zelda}
\end{table*}

\begin{table*}
\centering
\begin{tabular}{|ll|lllllllllll|}
\toprule
    Training set & Policy & \multicolumn{5}{l}{Score}
     & \multicolumn{5}{l}{Win rate} & avg. \\
 &  &  lvl0 &  lvl1 &  lvl2 &  lvl3 &  lvl4 &  lvl0 &  lvl1 &  lvl2 &  lvl3 &  lvl4 & win\\
\midrule
seaquest & A2C &  \cellcolor{gray!25}68(21) &  \cellcolor{gray!25}80(30) &  348(667) &  565(1249) &  411(992) &  \cellcolor{gray!25}0.88 &  \cellcolor{gray!25}0.82 &  0.70 &  0.15 &  0.10 & 0.53   \\
(lvl0, lvl1)          & GAP &  \cellcolor{gray!25}69(17) &  \cellcolor{gray!25}107(132) &  81(214) &  130(329) &  54(184) &  \cellcolor{gray!25}0.87 &  \cellcolor{gray!25}0.87 &  0.57 &  0.28 &  0.23 & 0.56  \\
  & PopArt &  \cellcolor{gray!25}57(222) &  \cellcolor{gray!25}262(609) &  1561(1550) &  358(997) &  1408(1912) &  \cellcolor{gray!25}0.00 &  \cellcolor{gray!25}0.15 &  0.47 &  0.02 &  0.12 & 0.15 \\
\midrule
seaquest0 & A2C &  \cellcolor{gray!25}774(574) &  793(464) &  2313(1264) &  105(311) &  804(1739) &  \cellcolor{gray!25}0.95 &  1.00 &  0.75 &  0.00 &  0.05 & 0.55  \\
(lvl0)          & GAP & \cellcolor{gray!25}\textbf{874}(697) & 529(527) & 0(1) & 0(0) & 100(308) & \cellcolor{gray!25}0.90 &  0.80 &  0.00 &  0.00 &  0.00 & 0.34 \\
  & PopArt &  \cellcolor{gray!25}7(9) &  57(226) &  1006(1526) &  54(225) &  404(823) &  \cellcolor{gray!25}0.00 &  0.10 &  0.20 &  0.00 &  0.00 & 0.06   \\
\midrule
seaquest2 & A2C &  \cellcolor{gray!25}550(502) &  833(668) &  2317(854) &  \cellcolor{gray!25}4117(1166) &  5333(1052) &  \cellcolor{gray!25}1.00 &  1.00 &  1.00 &  1.00 &  \cellcolor{gray!25}1.00 & 1.00  \\
(lvl0, lvl3) & GAP &  \cellcolor{gray!25}800(608) &  750(543) &  2525(1086) &  \cellcolor{gray!25}4050(1108) &  5350(1145) &  \cellcolor{gray!25}1.00 &  1.00 &  1.00 &  1.00 &  \cellcolor{gray!25}1.00 & 1.00 \\
  & PopArt &  \cellcolor{gray!25}494(831) &  \textbf{1400}(739) &  \textbf{4170}(947) &  \cellcolor{gray!25}\textbf{5822}(3174) &  \textbf{7955}(3075) &  \cellcolor{gray!25}0.22 &  0.81 &  1.00 &  0.79 &  \cellcolor{gray!25}0.88 & 0.74\\
\bottomrule
\end{tabular}
\caption{Evaluation results for \textit{Seaquest}.}
\label{tab:seaquest}
\end{table*}

\section{Conclusion and Future Work} \label{sec:conc}

In this work 3 versions of the A2C algorithm have been compared on 4 games with different sets of training levels. The trained agents then have been evaluated over all 5 levels available on the selected games and their ability of generalisation have been measured by their score in the game and their win rates. Our contribution is the comparison of these algorithms, the evaluation of the trained policies on the evaluation levels and the comparison of using different training sets, which capture different features of the game. We found that RL algorithms learn to play well on the training levels of the selected games, where almost all agents achieved high scores and near 100\% win rates, except on \texttt{zelda} which likely comes from the stochasticity of the game.


Training RL agents without evaluation levels, can be very deceptive as they do not show the robustness of the learned policies. Having robust policies is important, especially for games as players always find exploits in NPCs, which makes playing against them less enjoyable. 

Smartly selecting the training levels can improve generalisation, for example on seaquest, when lvl3 was present in the training set the agents learned to focus on collecting the divers on all levels. A negative example of the training levels is \texttt{zelda} in which lvl1 made the learning too difficult, that the agents failed to learn any good policy on any of the levels simultaneously. The objective of the RL algorithms is to maximize the discounted reward, not to win the game. Winning and getting a high score might correlate, but not in all cases, see the results of PopArt. Additionally, our results also showed that these algorithms struggle to learn in highly stochastic environments, for example in Zelda when both lvl0 and lvl1 were used for training. Too much stochasticity seem to cause a problem for the agent, while having deterministic games make the agent overfit to the training levels.

Recent works have shown the power of training on procedurally generated levels~\cite{cobbe2019leveraging}. A future line of work could be on identifying features in training levels, which could improve generalisation and robustness of the trained policies. Instead of randomly generating a large number of levels only a few could be enough if they capture enough features of the game.


\section{Acknowledgement}
This work was funded by the EPSRC Centre for Doctoral Training in Intelligent Games and Game Intelligence (IGGI) EP/L015846/1. This research utilised Queen Mary's Apocrita HPC facility (http://doi.org/10.5281/zenodo.438045),~supported by QMUL Research-IT.

\bibliography{references}
\bibliographystyle{unsrt}

\end{document}

